\documentclass[conference]{IEEEtran}
\IEEEoverridecommandlockouts

\usepackage{cite}
\usepackage{amsmath,amssymb,amsfonts}
\usepackage{algorithmic}
\usepackage{graphicx}
\usepackage{textcomp}
\usepackage{xcolor}

\usepackage{url}
\usepackage{subcaption}
\usepackage{multirow}
\usepackage{booktabs}
\usepackage{colortbl}
\usepackage{ascii}
\usepackage{paralist}
\usepackage{marvosym}

\usepackage[breaklinks]{hyperref}

\newcommand{\etal}{\textit{et al.}}

\renewcommand{\thefigure}{\Roman{figure}}
\renewcommand{\figurename}{FIGURE}

\def\BibTeX{{\rm B\kern-.05em{\sc i\kern-.025em b}\kern-.08em
    T\kern-.1667em\lower.7ex\hbox{E}\kern-.125emX}}

\makeatletter 
\newcommand{\linebreakand}{%
  \end{@IEEEauthorhalign}
  \hfill\mbox{}\par
  \mbox{}\hfill\begin{@IEEEauthorhalign}
}
\makeatother 

\begin{document}

\title{Large Language Model Aided\\ QoS Prediction for Service Recommendation
\thanks{This work is supported by the National Natural Science Foundation of China (No.62272001) and the Opening Foundation of State Key Laboratory of Cognitive Intelligence, iFLYTEK (COGOS-2023HE01).}
}


\author{
\IEEEauthorblockN{Huiying Liu}
\IEEEauthorblockA{\textit{School of Computer Science and Technology} \\
\textit{Anhui University}\\
Hefei, Anhui, China. \\
liuhuiying.ahu@hotmail.com}
\and
\IEEEauthorblockN{Zekun Zhang}
\IEEEauthorblockA{\textit{Department of Computer Science} \\
\textit{Stony Brook University}\\
Stony Brook, New York, USA. \\
zekzhang@cs.stonybrook.edu}
\and
\IEEEauthorblockN{Honghao Li}
\IEEEauthorblockA{\textit{School of Computer Science and Technology} \\
\textit{Anhui University}\\
Hefei, Anhui, China. \\
salmon1802li@gmail.com}
\and

\linebreakand

\IEEEauthorblockN{Qilin Wu}
\IEEEauthorblockA{\textit{School of Information Engineering} \\
\textit{Chaohu University}\\
Hefei, Anhui, China. \\
qlw@chu.edu.cn}
\and
\IEEEauthorblockN{Yiwen Zhang}
\IEEEauthorblockA{\textit{School of Computer Science and Technology} \\
\textit{Anhui University}\\
Hefei, Anhui, China. \\
zhangyiwen@ahu.edu.cn}
}

\maketitle

\begin{abstract}
Large language models (LLMs) have seen rapid improvement in the recent years, and have been used in a wider range of applications. After being trained on large text corpus, LLMs obtain the capability of extracting rich features from textual data. Such capability is potentially useful for the web service recommendation task, where the web users and services have intrinsic attributes that can be described using natural language sentences and are useful for recommendation. In this paper, we explore the possibility and practicality of using LLMs for web service recommendation. We propose the large language model aided QoS prediction (llmQoS) model, which use LLMs to extract useful information from attributes of web users and services via descriptive sentences. This information is then used in combination with the QoS values of historical interactions of users and services, to predict QoS values for any given user-service pair. On the WSDream dataset, llmQoS is shown to overcome the data sparsity issue inherent to the QoS prediction problem, and outperforms comparable baseline models consistently.
\end{abstract}

\begin{IEEEkeywords}
Service Recommendation, Quality of Service, Large Language Model.
\end{IEEEkeywords}

\section{Introduction}

\begin{figure*}[!ht]\centering
\includegraphics[width=0.7\linewidth]{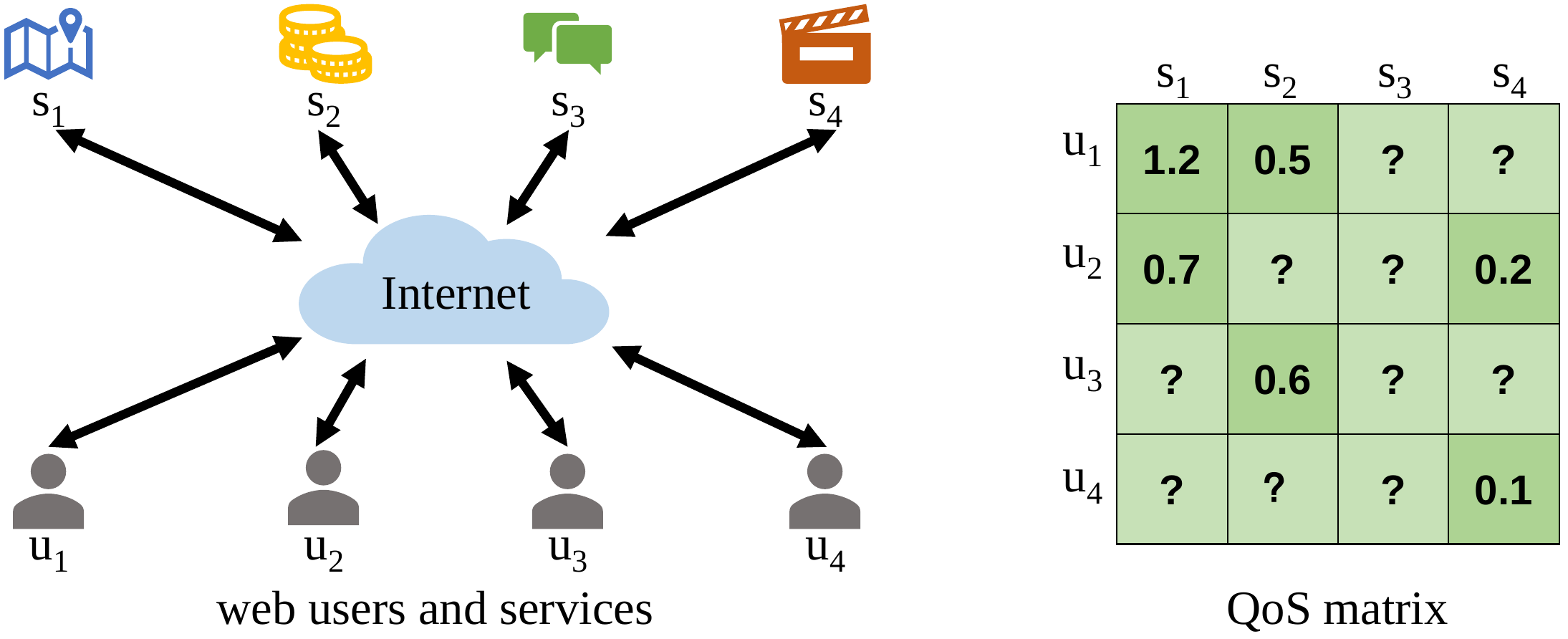}
\caption{Illustration of the QoS prediction problem. A QoS matrix exists for sets of web users and services, and each elements of the matrix corresponds to a user-service interaction. A QoS prediction model aims to predict the value of unknown QoS values from the known QoS values from historical user-service interactions.}\label{fig:intro-QoS}
\end{figure*}

The rapid development of cloud computing has led to an explosive growth in the number of cloud services, resulting in many services having identical or similar functionalities \cite{introduction1}. There is a growing need for service providers to provide the most suitable service for users to meet customer demands. When there exist many candidate services for a user request, quality of service (QoS) is a commonly used metric to measure the performance of these services. QoS refers to non-functionality aspects of services \cite{introduction2, SSE2, SSE5} such as throughput, response time, and cost. It plays a crucial role in distinguishing similar services and helps users select the most appropriate one. However, in real-world scenarios, evaluating the QoS of a large number of services on the client side is expensive and time-consuming due to the sheer scale. Additionally, most end users lack the expertise to do the evaluation \cite{introduction3}. Therefore, it is common for an end user is to predict the QoS values for an unknown service, and the QoS values are used to help the user find the most suitable service.

The problem of QoS prediction is illustrated in \autoref{fig:intro-QoS}. For the sets of web users (u$_1$ $\sim$ u$_4$) and services (s$_1$ $\sim$ s$_4$) exists a QoS matrix, in which each value corresponds to a user-service interaction. The matrix contains some known QoS values from historical user-service interactions, and our goal is to predict unknown QoS values (marked by ``?'') using these known values, thereby recommending more suitable services to users. Many methods have been developed to predict the QoS values for a given user and service. Among them, collaborative filtering is the most widely used technique. It aim to accurately estimate unknown QoS values by learning from the QoS value matrix from historical user-service interactions. However, in practice, the historical QoS value matrix is often highly sparse \cite{SSE4} , as each user usually only interacted with a limited number of cloud services in the past. This data sparsity issue poses difficulty for many QoS prediction methods. Although many methods aim to mitigate this difficulty, it is still challenging to achieve ideal predictive performance.

More recently, large language models (LLMs) \cite{Zhao2023ASO, Minaee2024LargeLM} have demonstrated remarkable performance in understanding and generating texts. LLMs have large capacity and are trained on vast amounts of data, enabling them to capture complex patterns in natural language with a long context. They have rapidly become powerful tools for various natural language processing tasks \cite{devlin2019bert, liu2019roberta, brown2020languagemodelsfewshotlearners, abdin2024phi3, meta2024llama3}. For the QoS prediction problem, each user and service has some specific attributes. For instance, users may have attributes such as its country of origin and the autonomous system it belongs to. Given that LLMs are capable of handling natural language with contextual information, we claim that such capability can be used for QoS prediction task as well. In this paper, we propose to use natural language sentence to describe each user or service using their corresponding attributes. Those attributes contain information such as the proximity and type, which are potentially useful for QoS prediction. We then use state-of-the-art LLMs to extract the descriptive features of users and services from those sentences. This LLMs-based feature helps the QoS prediction model better understand and represent the latent features of users and services. It effectively mitigates the data sparsity issue, as the descriptive LLM features provide information that is complementary to the historical user-service interactions. Along with the historical QoS values, it helps the model to provide more reliable and effective service recommendations.

We summarize our main contributions as the following:

\begin{compactitem} 
\item To the best of our knowledge, we first introduce large language models for the web service recommendation task. We propose large language model aided QoS prediction (\textbf{llmQoS}), a novel approach that combines collaborative filtering and nature language processing, making use of both historical user-service interaction information and rich text feature from attributes of users and services. It can effectively mitigate the data sparsity issue inherent to the QoS prediction problem.

\item The proposed llmQoS model is robust and versatile. It can work with different LLMs, and can generalize to different collaborative filtering network architectures.

\item We show that llmQoS can predict QoS values accurately under different data sparsity levels. On the WSDream dataset, it outperforms several existing QoS prediction models consistently.

\end{compactitem}

\section{Related Work}

\subsection{QoS Prediction for Service Recommendation}

Among QoS prediction methods for web services, collaborative filtering (CF) \cite{CF} methods receive extensive attention and are widely regarded as the most effective approach. CF-based QoS prediction methods can be divided into two main categories: memory-based methods and model-based methods.

For memory-based CF methods, the similarity of users or services is first calculated. Then QoS values of the target user and service are predicted based on both the historical QoS values and the similarity relationships with their neighbors. Shao \etal{} \cite{user-based1} first utilize user-based CF methods for QoS prediction. Qi \etal{} \cite{user-based-cold} identify ``enemies'' of the target user to determine their ``potential friends'', successfully overcoming the cold-start problem. Meanwhile, item-based methods are widely employed \cite{item-based1, item-based2}. Chen \etal{} \cite{item-based-loc} propose a predictive method that integrates service similarity and geographical location information, effectively alleviating data sparsity and cold-start issues. Additionally, hybrid algorithms that compute both service-to-service and user-to-user similarities \cite{item-user-based1} are developed. Zheng \etal{} \cite{UIPCC} simultaneously consider user and service similarities for QoS prediction, demonstrating superior predictive performance compared to previous methods. Hu \etal{} \cite{time_and_sim} propose to integrate time information to improved CF for QoS prediction.

Model-based CF methods typically use models to predict the QoS values, with matrix factorization (MF) being the most common approach. MF methods decompose the user-service interaction matrix into user and service matrices, predicting values by multiplying these two matrices. Zhu \etal{} \cite{AMF} notably extend the traditional MF models by incorporating techniques including data transformation, online learning, and adaptive weighting on the basis of CF techniques. Zheng \etal{} \cite{NIMF} develop a neighborhood-integrated MF method for personalized web service QoS prediction. Xu \etal{} \cite{PMF-QoS} employ a probabilistic matrix factorization (PMF) model to learn and predict QoS values, incorporating geographical location to identify user neighbors and comprehensively consider their impact on web service invocation experiences.

Deep learning technology has demonstrated remarkable capabilities across various fields of artificial intelligence in recent years \cite{Silver2016AphaGo, Chai2021DLCV, Soori2023DLRobotics, Zhao2023ASO, SSE1, Minaee2024LargeLM}. And an increasing number of deep learning based techniques for QoS prediction are developed \cite{deeplearning1, deeplearning2, SSE6} too. Zou \etal{} \cite{NCRL} propose a novel adaptive QoS prediction framework, utilizing a dual-tower deep residual network to extract latent features of users and services and predict QoS values. Wu \etal{} \cite{Robust} introduce a reputation-integrated graph convolutional network for robust and accurate QoS prediction. Wei \etal{} \cite{DRGL} use a transformer based method to learn the graph structure for time-aware service recommendation. Zhu \etal{} \cite{BGCL} develop a twin graph network based on graph contrastive learning, effectively addressing cold-start and data sparsity issues. Zhou \etal{} \cite{SCATSF} propose the spatial context-aware time series forecasting (SCATSF) framework by utilizing both temporal and spatial contextual information of users and services. Lian \etal{} \cite{PMT} facilitate multi-task learning for training QoS prediction model.

Most deep learning based QoS prediction models use the user and service ID information through embedding layers. Additionally, to improve prediction performance, some methods \cite{NCRL, BGCL} utilize more information of the users and services such as their location, IP address, and the autonomous systems they belong to. Such information provides similarity or proximity signal for the model, which are useful for QoS prediction. In our work, we utilize such information through large language models.

\subsection{Large Language Model}\label{sec:related-llm}

In the field of natural language processing (NLP), LLMs have seen rapid development since the introduction of the transformer architecture \cite{NIPS2017Transformer}. They have shown state-of-the-art performance across virtually all NLP tasks, including text classification and summarizing, machine translation, text generation, and human-like conversation systems \cite{Zhao2023ASO, Minaee2024LargeLM}.

Most LLMs are based on variants of the already ubiquitous transformer architecture, with the core components being the attention based mechanism that can effectively capture the correlation among different words in the text. It is common for a group of researchers to develop a series of LLMs that share similar architecture and training recipes, but with generational improvement in performance and capability. Notable ones include the BERT series of original BERT\cite{devlin2019bert}, RoBERTa \cite{liu2019roberta}, ALBERT \cite{lan2020albert}, and DeBERTa \cite{liu2019roberta}; the OpenAI GPT series of GPT \cite{Radford2018ImprovingLU}, GPT-2 \cite{solaiman2019release}, GPT-3 \cite{brown2020languagemodelsfewshotlearners}, and GPT-4 \cite{openai2024gpt4technicalreport}; the META LLaMA series of LLaMA \cite{Touvron2023LLaMAOA}, LLaMA2 \cite{Touvron2023Llama2O}, and LLaMA3 \cite{meta2024llama3}; and the Microsoft Phi series of Phi1 \cite{gunasekar2023textbooksneed, li2023textbooksneediiphi15}, Phi2 \cite{ms2023phi2}, and Phi3 \cite{abdin2024phi3}. The common trend in each series is that the newer models usually have more parameters, and are trained on larger corpus of text. Larger LLMs have higher capacity, can learn more information from a larger corpus of text, have a higher generalization capability, and in general perform better on standardized NLP benchmarks \cite{HFOpenLLMLeaderboard}.

Thought LLMs are trained from textual data, they can generalize to tasks beyond NLP too. They have been shown to work on visual question-answering \cite{Guo2022FromIT}, open-vocabulary object detection  \cite{liu2023grounding}, planning \cite{song2023llmplanner}, and autonomous driving \cite{fu2023drive}. In this paper, we use LLMs for the service recommendation task. When using LLMs on non-NLP tasks, it is important to convert the data into a form of prompt that can be processed by the LLMs. In our work we construct descriptive sentences from the relative attributes of web users and services to be used by LLMs. More details are discussed in \autoref{sec:llm-extraction}.

Different LLMs have different levels of accessibility. The parameters of some LLMs can be downloaded and ran on local computers for any end user, which is often referred as ``open-weight'' models. Some models are only be used through API calls and the model parameters cannot be accessed. In this paper we only use open-weight models for reproducibility. However our method can be applied to any LLMs with feature extraction functionality.

\section{Methods}

\subsection{Large Language Model Feature Extraction}\label{sec:llm-extraction}

\begin{figure*}[!ht]\centering
\includegraphics[width=0.9\linewidth]{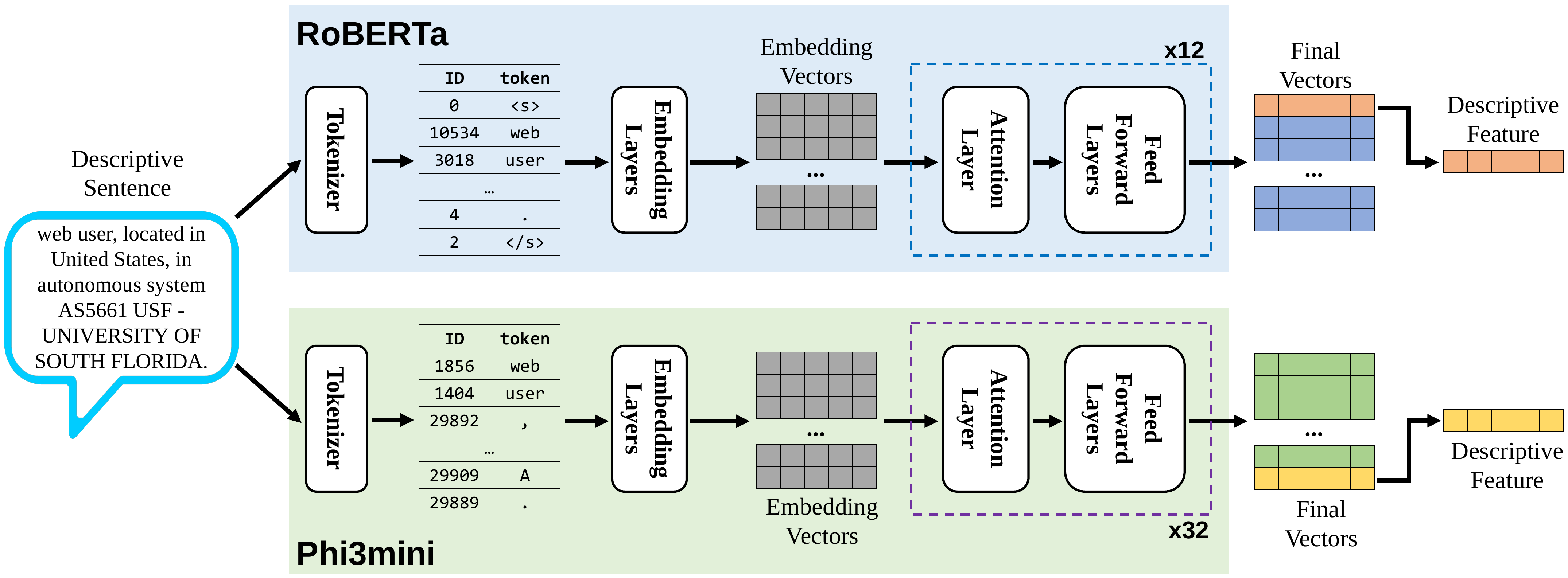}
\caption{The proposed llmQoS model uses large language models to extract text features from descriptive sentences of user or service. The descriptive sentence is constructed from certain attributes for each user or service, then converted to a sentences-level feature vector, which is used for QoS prediction. Please refer to \autoref{sec:llm-extraction} for more details.}\label{fig:llm-extraction}
\end{figure*}

For each web user and service in WSDream dataset, we use the corresponding attributes to construct a descriptive sentence as the input to the LLMs for feature extraction. Specifically, we use the country and autonomous system for users. For instance, the descriptive sentence for one of the user is {\asciifamily "web user, located in United States, in autonomous system AS5661 USF - UNIVERSITY OF SOUTH FLORIDA."} Similarly, the URL, service provider, country, and autonomous system are used to services. For example one of the services has the descriptive sentence of {\asciifamily "web service, at url http://biomoby.org/services/wsdl/ualberta.ca /DrugBankByName, hosted by ualberta.ca, located in Canada, in autonomous system AS3359 University of Alberta."} We do not include numerical attributes such as IP address, latitude, and longitude in the sentences, as LLMs usually cannot extract meaningful information from numbers. For each user or service, we use the same descriptive sentence for different LLMs.

After obtaining the descriptive sentence for each user or service, we use pre-trained RoBERTa \cite{liu2019roberta} or Phi3mini \cite{abdin2024phi3} to extract sentence-level language features. The pipeline is illustrated in \autoref{fig:llm-extraction}. The sentence is first converted into a sequence of tokens by a tokenizer. The embedding vector of each token is obtained by their token ID and position. A set of stacked LLM layers process on this sequence of embedding vectors to obtain the final sequence of feature vectors. One of the vector in the final sequence is used as the LLM feature vector of the descriptive sentence.

RoBERTa and Phi3mini have similar overall architecture. Each LLM layer in them comparably consists of an attention based layer \cite{NIPS2017Transformer} and a set of feed-forward layers. The attention layer calculates the correlation among the feature vectors in the sequence. The feed-forward layers are combination of linear layers, normalization layers, and activation functions. This common architecture is used by many other LLMs too.

However, RoBERTa and Phi3mini have different number of layers, different dimensionality of the embedding vectors, and different tokenizer. Phi3mini has $3.5\times 10^9$ parameters, which is significantly larger than RoBERTa's $1.2\times 10^8$ million parameters. They are also pre-trained on different datasets using different training tasks. RoBERTa is pre-trained to do text classification on total of 160 giga-bytes of text data. If each token uses 5 bytes on average, the model sees $3.2\times 10^{10}$ tokens during training. For text classification, the first vector in the final sequence is used. So we also use the first vector as the sentence-level feature. On the other hand, Phi3mini is pre-trained using next token prediction on a dataset with $3.3\times 10^{12}$ tokens. Next token prediction uses the last vector in the final sequence, so we also use the last vector in the final sequence for Phi3mini. Later we show in experiments that Phi3mini outperforms RoBERTa consistently.

\subsection{Model Architecture}\label{sec:model-arch}

\begin{figure*}[!ht]\centering
\includegraphics[width=0.9\linewidth]{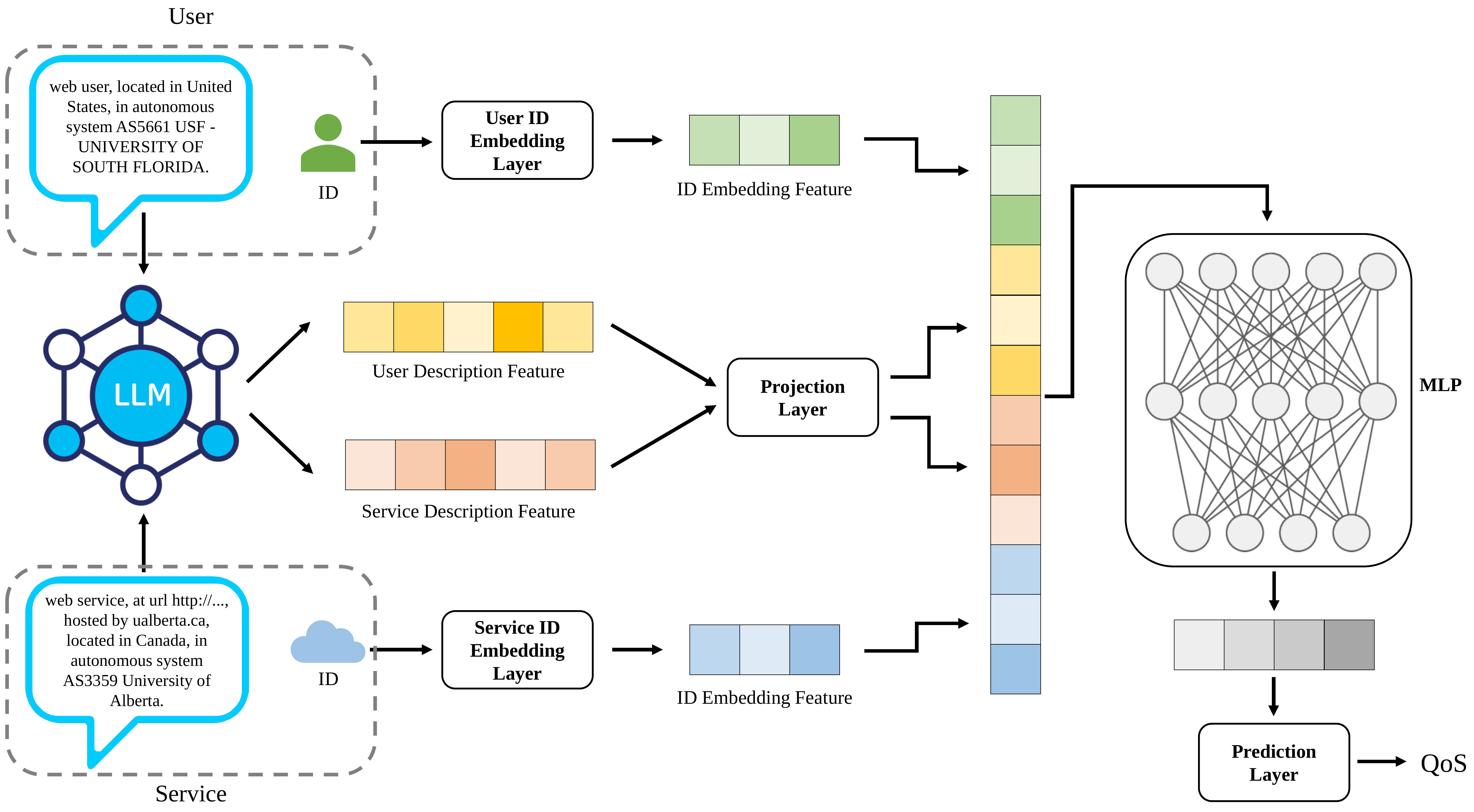}
\caption{The overall architecture of the proposed llmQoS model. The embedding features for user and service are combined with the descriptive LLM features described in \autoref{sec:llm-extraction}, and used by an MLP network and a prediction layer to get the final prediction of the QoS value. Please refer to \autoref{sec:model-arch} for more details.}\label{fig:model-architecture}
\end{figure*}

The overall architecture of the proposed large language model aided QoS prediction (llmQoS) model is illustrated in \autoref{fig:model-architecture}. First, both the user and service IDs are projected to feature vectors by their corresponding embedding layers. Formally for user $u$ and service $s$,
\begin{equation}
\mathbf{e}_u = Embed_u (i_u),
\end{equation}
\begin{equation}
\mathbf{e}_s = Embed_s (i_s),
\end{equation}
where $i_u$ and $i_s$ are the IDs of the user and the service, and $Embed_u$ and $Embed_s$ are the corresponding embedding layers.

Then descriptive sentences are constructed and the LLM features are extracted for each user and service based on their attributes, as described in \autoref{sec:llm-extraction}. The descriptive LLM features have much larger dimensionality compared to the ID embedding features. So a projection layer is used for to reduce them to the same dimensionality as the ID embedding features:
\begin{equation}
\mathbf{f}_u = Proj (LLM (t_u)),
\end{equation}
\begin{equation}
\mathbf{f}_s = Proj (LLM (t_s)),
\end{equation}
where $t_u$ and $t_s$ are the descriptive sentences of the user and service. Please note that we use the same $LLM$ model and projection layer $Proj$ for both user and service.

Then the user ID embedding feature vector, the projected user descriptive LLM feature vector, the service ID embedding feature vector, and the projected service descriptive LLM feature vector are all concatenated. This concatenated feature vector is used as the input to an multi-layer perceptron (MLP) network. Finally a linear layer produces the predicted QoS value for the input user-service pair. Formally,
\begin{equation}
\hat{y} = Linear (MLP ([\mathbf{e}_u; \mathbf{f}_u; \mathbf{e}_s; \mathbf{f}_s])),
\end{equation}
where $\hat{y}$ is the predicted QoS value for $u$ and $s$.

Please note that our network design is straightforward without complicated architectures such as graph convolution or attention. The QoS prediction network architecture is not the focus of this paper. With straightforward architecture, llmQoS can still achieve state-of-the-art performance as shown by the experiments. This demonstrate the effectiveness of incorporating LLM features for QoS prediction. In \autoref{sec:bgcl} we further demonstrate that LLM features can be integrated into more complex QoS prediction models and can still increase the performance significantly and consistently.

\section{Experiments}

This section shows the results and discussion of our extensive experiments with llmQoS. It starts with the details of the dataset, evaluation metrics, implementation, and the baseline methods. Then the performance of llmQoS is compared with the baselines. And multiple ablation study experiments are conducted to verify that the LLM feature is beneficial for QoS prediction, and can be integrated to other network architectures too.


\begin{table*}[!ht]\centering
\setlength{\tabcolsep}{10pt}
\caption{The QoS prediction performance in terms of Mean Average Error (MAE) and Root Mean Squared Error (RMSE) of different baseline models and the proposed llmQoS model. Phi3mini is used for llmQoS in this experiment. The best baseline method is marked by \colorbox{lightgray}{gray background}, and we show the percentage improvement from the best baseline achieved by llmQoS (Gain). Our proposed model outperforms all baselines at all densities consistently.}\label{tab:results}
\begin{tabular}{ccccccccc}\toprule
\multirow{3}{*}{Model}&\multicolumn{8}{c}{Sub-dataset throughput}\\\cmidrule(lr){2-9}
& \multicolumn{2}{c}{$D$=5\%} & \multicolumn{2}{c}{$D$=10\%} & \multicolumn{2}{c}{$D$=15\%} & \multicolumn{2}{c}{$D$=20\%} \\
\cmidrule(lr){2-3}\cmidrule(lr){4-5}\cmidrule(lr){6-7}\cmidrule(lr){8-9}
& MAE & RMSE & MAE & RMSE & MAE & RMSE & MAE & RMSE \\\midrule
UIPCC & 26.757 & 60.799 &22.370 &54.456 &20.219 &50.704 &18.928 &48.295 \\
RegionKNN & 25.632 &67.868 &24.838 &67.551 &24.584 &67.314 &24.036 &66.176 \\
LACF & 23.169 &58.967 &19.626 &53.105 &17.795 &49.766 &16.667 & 47.625 \\
PMF & 19.082 &57.883 &15.994 &48.071 &14.670 &44.013 &13.924 
&41.714  \\
BGCL & 20.655 &61.297 &19.318 &59.134 &18.134 &58.804 &18.017 
&58.689 \\
PSO-USRec &23.332 &60.155 &19.740 &54.254 &17.839 &50.209 &16.787 &47.395 \\
LMF-PP & 18.301 &51.777 &15.913 & \cellcolor[gray]{.8}46.142 &14.745 &\cellcolor[gray]{.8}42.993 &14.103 & \cellcolor[gray]{.8}41.408 \\
DCALF & \cellcolor[gray]{.8}17.658 & \cellcolor[gray]{.8}51.632& \cellcolor[gray]{.8}15.360&46.428 & \cellcolor[gray]{.8}14.384 &43.402& \cellcolor[gray]{.8}13.670 &41.624  \\\midrule
\multirow{1}{*}{llmQoS}&\textbf{13.714} &\textbf{46.635}  &\textbf{12.022} &\textbf{42.947} &\textbf{11.156}  &\textbf{39.567} &\textbf{10.760} &\textbf{38.365}  \\
\multirow{1}{*}{Gain} & 22.33\% &9.68\%&21.73\%&6.92\%&28.93\%&7.97\%&21.28\%&7.35\%  \\\midrule\midrule
\multirow{3}{*}{Model}&\multicolumn{8}{c}{Sub-dataset response time}\\\cmidrule(lr){2-9}
& \multicolumn{2}{c}{$D$=5\%} & \multicolumn{2}{c}{$D$=10\%} & \multicolumn{2}{c}{$D$=15\%} & \multicolumn{2}{c}{$D$=20\%} \\
\cmidrule(lr){2-3}\cmidrule(lr){4-5}\cmidrule(lr){6-7}\cmidrule(lr){8-9}
& MAE & RMSE & MAE & RMSE & MAE & RMSE & MAE & RMSE \\\midrule
UIPCC & 0.625 & 1.388 &0.582 &1.330 &0.501 &1.250 &0.450 &1.197  \\
RegionKNN &0.588 &1.543 &0.548 &1.513 &0.526 &1.513 &0.516 &1.521 \\
LACF &0.637 &1.444 &0.566 &1.342 &0.516 &1.276 &0.483 &1.230 \\
PMF &0.569 &1.537 &0.487 &1.316 &0.452 &1.221 &0.431 &1.169 \\
BGCL &\cellcolor[gray]{.8}0.461 &1.407 &\cellcolor[gray]{.8}0.433&1.374 &\cellcolor[gray]{.8}0.424&1.334 &\cellcolor[gray]{.8}0.416&1.320 \\
PSO-USRec &0.565 &1.358 &0.506 &1.274 &0.471 &1.222 &0.444 &1.181 \\
LMF-PP &0.529 & \cellcolor[gray]{.8}1.341 &0.473 & \cellcolor[gray]{.8}1.242 &0.447 &\cellcolor[gray]{.8}1.210 &0.426 & \cellcolor[gray]{.8}1.161 \\
DCALF &0.546 &1.402 &0.486 &1.265 &0.464 &\cellcolor[gray]{.8}1.210&0.452 &1.176 \\\midrule
\multirow{1}{*}{llmQoS} &\textbf{0.409} &\textbf{1.290} &\textbf{0.360} &\textbf{1.224} &\textbf{0.344} &\textbf{1.186} &\textbf{0.327} &\textbf{1.159}  \\
\multirow{1}{*}{Gain}&11.36\% &3.84\%&16.72\%&1.41\%&18.93\%&2.01\%&21.40\%&0.19\%  \\\bottomrule
\end{tabular}
\end{table*}

\begin{figure*}[!ht]\centering
\begin{subfigure}{0.24\linewidth}
    \caption{Throughput $D$=5\%}
    \includegraphics[width=\linewidth]{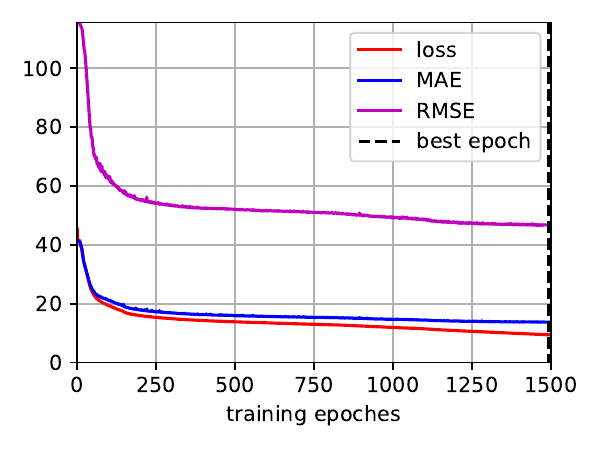}
\end{subfigure}\hfill
\begin{subfigure}{0.24\linewidth}
    \caption{Throughput $D$=10\%}
    \includegraphics[width=\linewidth]{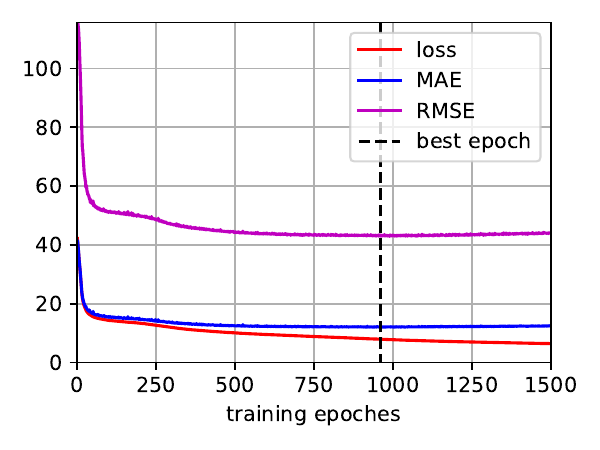}
\end{subfigure}\hfill
\begin{subfigure}{0.24\linewidth}
    \caption{Throughput $D$=15\%}
    \includegraphics[width=\linewidth]{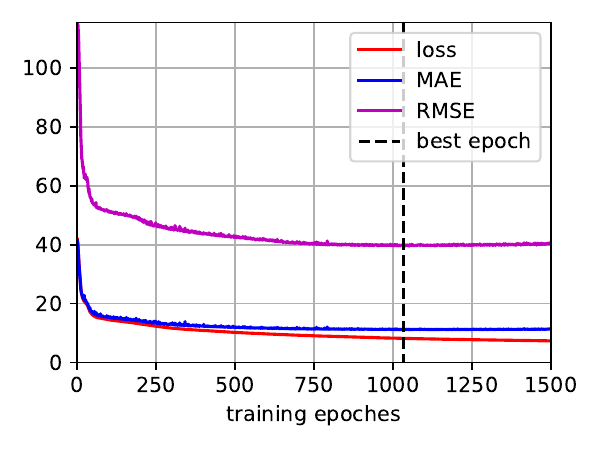}
\end{subfigure}\hfill
\begin{subfigure}{0.24\linewidth}
    \caption{Throughput $D$=20\%}
    \includegraphics[width=\linewidth]{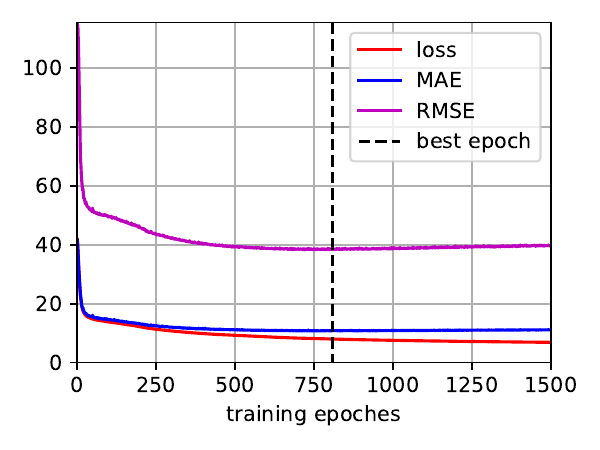}
\end{subfigure}
\begin{subfigure}{0.24\linewidth}
    \caption{Response time $D$=5\%}
    \includegraphics[width=\linewidth]{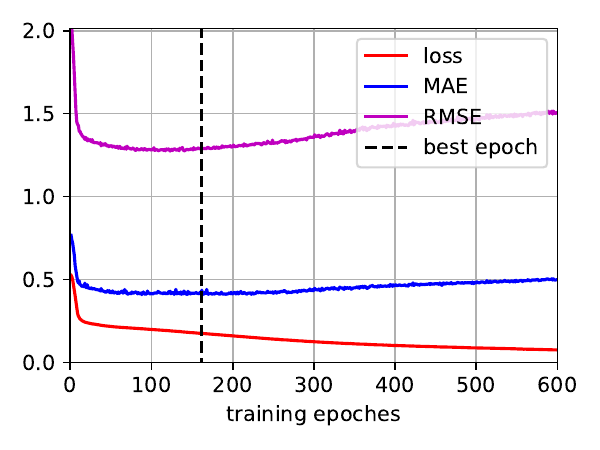}
\end{subfigure}\hfill
\begin{subfigure}{0.24\linewidth}
    \caption{Response time $D$=10\%}
    \includegraphics[width=\linewidth]{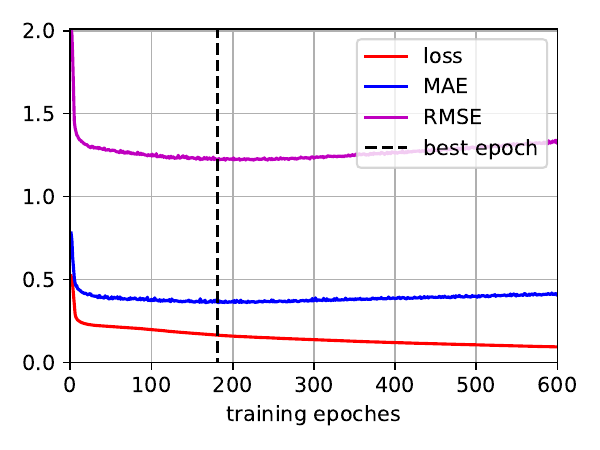}
\end{subfigure}\hfill
\begin{subfigure}{0.24\linewidth}
    \caption{Response time $D$=15\%}
    \includegraphics[width=\linewidth]{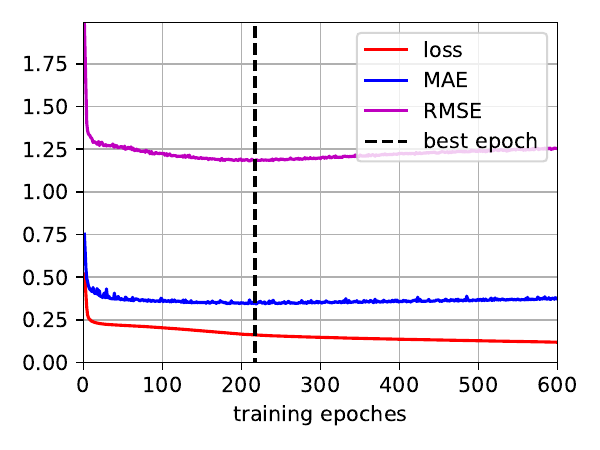}
\end{subfigure}\hfill
\begin{subfigure}{0.24\linewidth}
    \caption{Response time $D$=20\%}
    \includegraphics[width=\linewidth]{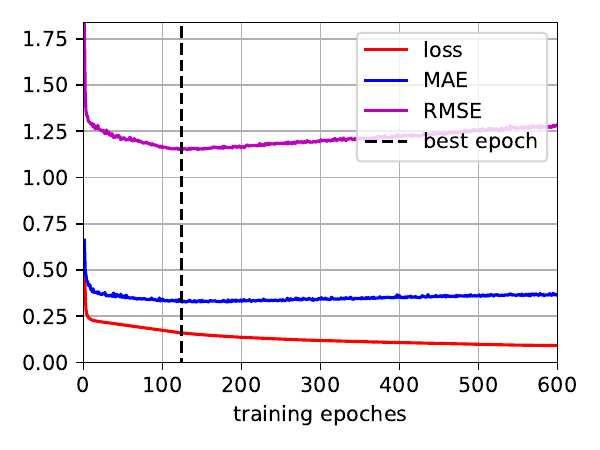}
\end{subfigure}
\caption{Curves of loss, MAE, and RMSE during the training of llmQoS. The loss continues decreasing while MAE and RMSE on testing set reach minimum earlier. The training epoch when the model achieves the lowest MAE is used as the final model.}
\label{fig:curve}
\end{figure*}

\subsection{Dataset}

To evaluate the performance of our model, we conducted extensive experiments on the WSDream dataset \cite{wsdream2008}. 
WSDream is a large-scale real-world dataset of Internet user-service interactions, and is widely recognized as a standard benchmark for evaluating QoS prediction models in this research domain, involving 5,825 services and 339 users in total. Each user and service is identified by a unique ID number. In addition to the IDs, WSDream also provides basic attribute information for each user and service. User attributes consist of IP Address, Country, IP Number, Autonomous System, Latitude, and Longitude. Service attributes include WSDL Address, Service Provider, IP Address, Country, IP Number, Autonomous System, Latitude, and Longitude. We utilize both the IDs and attributes for the users and services in our llmQoS model.

WSDream contains QoS record that include 1,873,850 response time values and 1,831,265 throughput values. For experiments, both the throughput sub-dataset and the response time sub-dataset are randomly split into training and testing sets. We use different density values $D$, which is the percentage of training set data in the total dataset, to test the models' prediction ability at different data sparsity levels. Specifically we use $D$=5\%, 10\%, 15\%, and 20\%. It is expected that with all other settings unchanged, QoS prediction models achieve higher precision with larger values of $D$.

\subsection{Implementation Details}

We implemented our model using {\asciifamily Keras 2.0.9} \cite{chollet2015keras} with a {\asciifamily TensorFlow 1.4.0} \cite{tensorflow2015whitepaper} backend. The ID embedding for both user and service has a dimensionality of 16. Both pre-trained RoBERTa and Phi3mini are taken from Huggingface \cite{huggingface}, and the descriptive LLM feature extraction is implemented using the {\asciifamily Transformers} library. The descriptive LLM feature vectors from RoBERTa and Phi3mini have 768 and 3,072 dimensions, respectively. In both cases the descriptive LLM feature vector is reduced to 16 dimensions by the projection layer. The MLP module has 3 linear layers with output dimensionalities of 32, 16, and 8. Each linear layer is followed by a ReLU activation function \cite{agarap2018deep}. The final prediction layer is a single linear layer that maps the 8-dimensional output vector of MLP module to a single QoS value.

For all experiments, we use Huber loss \cite{Huber1964RobustEO} to train the model. The loss function is defined as
\begin{equation}
\mathcal{L}(y, \hat{y}) = %
\begin{cases}
  \frac{1}{2} (y - \hat{y})^2, & \text{if } |y - \hat{y}| < \delta \text{;}\\
    \delta (|y - \hat{y}| - \frac{\delta}{2}), & \text{otherwise},
\end{cases}
\end{equation}
where $y$ is the ground truth QoS value, and $\hat{y}$ is the QoS value predicted by the model. We use threshold $\delta = 1.0$. All models are trained using the Adam optimizer \cite{KingmaB14Adam} with learning rate of $10^{-4}$ and batch size of 256.

We measure the Mean Average Error (MAE) and Root Mean Squared Error (RMSE) on the testing set to measure the performance of the model. They are defined as
\begin{equation}
\text{MAE} = \frac{\sum_{i=1}^{N} |y_i - \hat{y}_i|}{N},
\end{equation}
and
\begin{equation}
\text{RMSE} = \sqrt{\frac{\sum_{i=1}^{N} (y_i - \hat{y}_i)^2}{N}},
\end{equation}
where $(y_i, \hat{y}_i)$ is a pair of ground truth and predicted QoS values, and $N$ is the number of data points in the testing set. We measure the MAE of the model at the end of each training epoch, and use the one with lowest MAE as the final trained model. We observed that for throughput experiments, the final model can be obtained before 1,500 epochs. For response time experiments, the final model can be obtained before 600 epochs.

\subsection{Baseline Models}


We benchmark llmQoS against several diverse QoS prediction baselines, which span a range of methodologies from traditional approaches like classic collaborative filtering and matrix factorization, to location-aware techniques, and more recent graph-based or optimization methods. All baselines use identical experimental settings for fair comparison.

\noindent\textbf{UIPCC} \cite{UIPCC} is a classic collaborative filtering method predicting QoS from historical user-service invocation data.

\noindent\textbf{RegionKNN} \cite{RegionKNN} is a hybrid collaborative filtering algorithm that considers geographic information for service recommendations.


\noindent\textbf{LACF} \cite{LACF} is a location-aware collaborative filtering method using integrated user and service location data for QoS prediction.


\noindent\textbf{PMF} \cite{PMF} is a probabilistic matrix factorization technique applied to QoS prediction.

\noindent\textbf{BGCL} \cite{BGCL} is a graph-based contrastive learning method that utilizes graph attention mechanisms to incorporate domain information of users and services for QoS prediction.


\noindent\textbf{PSO-USRec} \cite{PSO-USRec} is an optimization-based QoS prediction model using particle swarm optimization to search the global QoS distribution space and refine predictions.


\noindent\textbf{LMF-PP} \cite{LMF-PP} is a location-based matrix factorization method integrating invocation/domain information and fused similarity for preference propagation in QoS prediction.


\noindent\textbf{DCALF} \cite{DCALF} is a method to detect domains and noise in the QoS matrix via density peak-based clustering for QoS prediction.

\begin{figure*}[!ht]\centering
\begin{subfigure}{0.24\linewidth}
    \caption{Throughput $D$=5\%}
    \includegraphics[width=\linewidth]{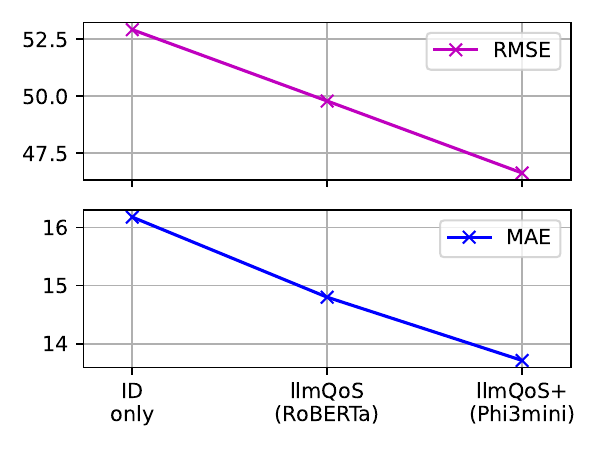}
\end{subfigure}\hfill
\begin{subfigure}{0.24\linewidth}
    \caption{Throughput $D$=10\%}
    \includegraphics[width=\linewidth]{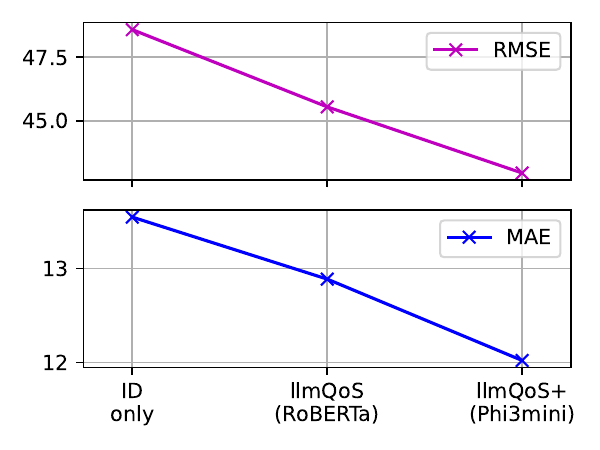}
\end{subfigure}\hfill
\begin{subfigure}{0.24\linewidth}
    \caption{Throughput $D$=15\%}
    \includegraphics[width=\linewidth]{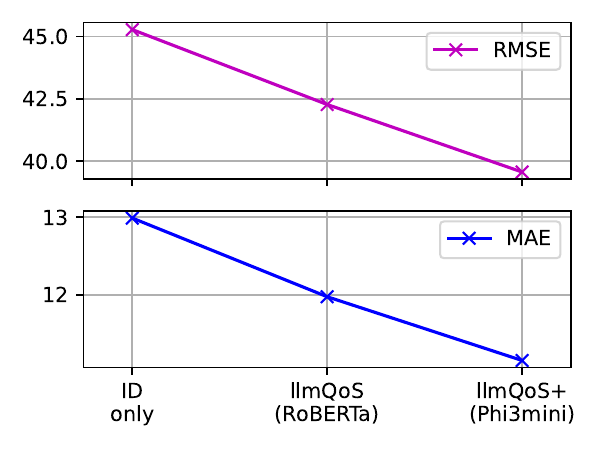}
\end{subfigure}\hfill
\begin{subfigure}{0.24\linewidth}
    \caption{Throughput $D$=20\%}
    \includegraphics[width=\linewidth]{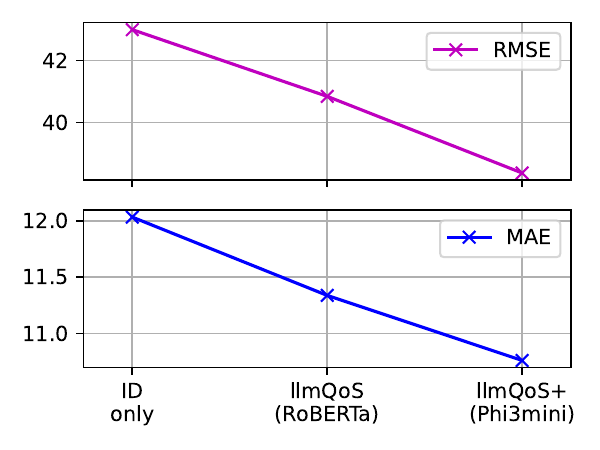}
\end{subfigure}
\begin{subfigure}{0.24\linewidth}
    \caption{Response time $D$=5\%}
    \includegraphics[width=\linewidth]{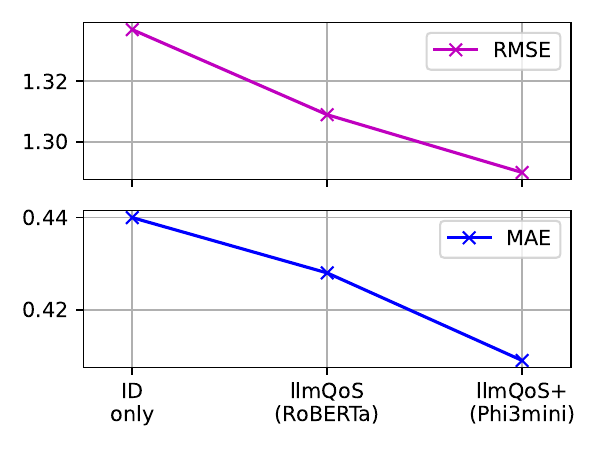}
\end{subfigure}\hfill
\begin{subfigure}{0.24\linewidth}
    \caption{Response time $D$=10\%}
    \includegraphics[width=\linewidth]{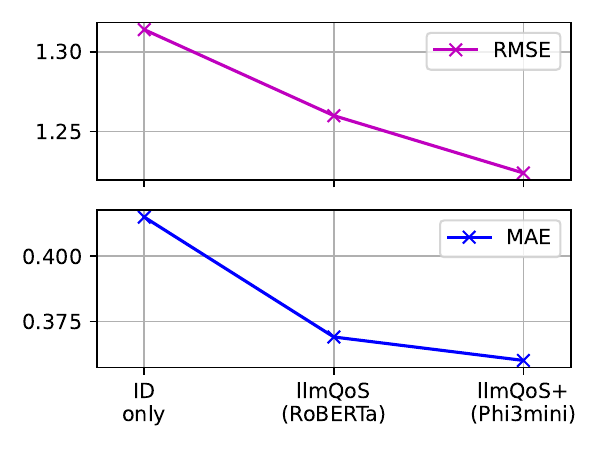}
\end{subfigure}\hfill
\begin{subfigure}{0.24\linewidth}
    \caption{Response time $D$=15\%}
    \includegraphics[width=\linewidth]{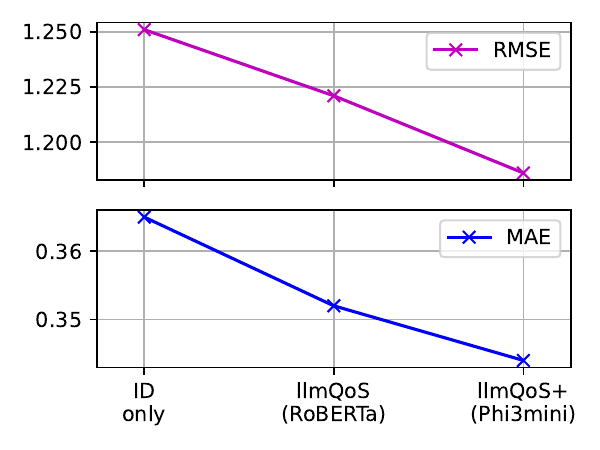}
\end{subfigure}\hfill
\begin{subfigure}{0.24\linewidth}
    \caption{Response time $D$=20\%}
    \includegraphics[width=\linewidth]{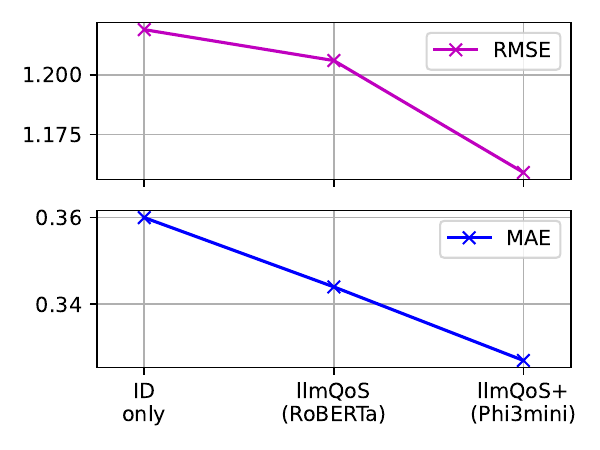}
\end{subfigure}
\caption{QoS prediction performance in terms of Mean Average Error (MAE) and Root Mean Squared Error (RMSE) of llmQoS with different LLM feature extractors. We compare Phi3mini with RoBERTa. Additionally a model without any LLM feature (ID only) is compared. Both LLM based methods outperform ID only model, while larger Phi3mini achieves better QoS prediction performance than RoBERTa.}
\label{fig:llm-ablation}
\end{figure*}

\begin{figure*}[!ht]\centering
\begin{subfigure}{0.24\linewidth}
    \caption{Throughput $D$=5\%}
    \includegraphics[width=\linewidth]{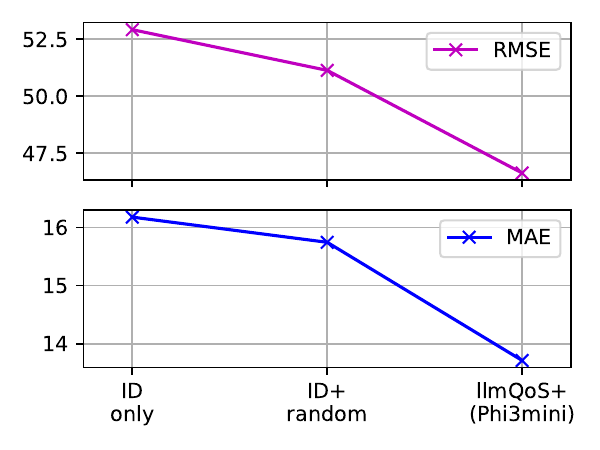}
\end{subfigure}\hfill
\begin{subfigure}{0.24\linewidth}
    \caption{Throughput $D$=10\%}
    \includegraphics[width=\linewidth]{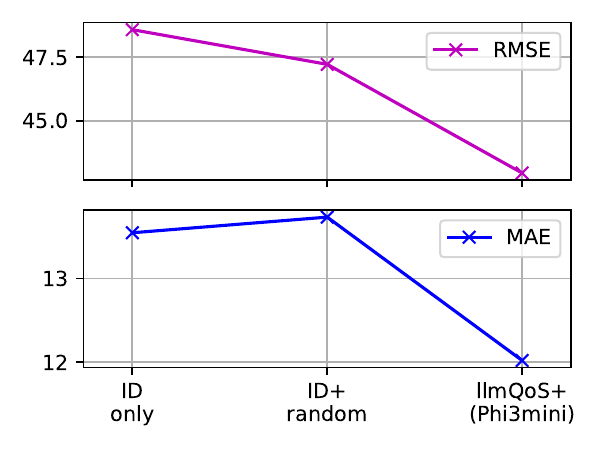}
\end{subfigure}\hfill
\begin{subfigure}{0.24\linewidth}
    \caption{Throughput $D$=15\%}
    \includegraphics[width=\linewidth]{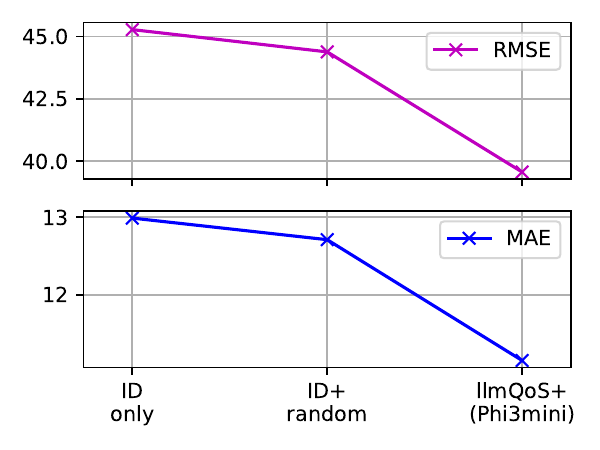}
\end{subfigure}\hfill
\begin{subfigure}{0.24\linewidth}
    \caption{Throughput $D$=20\%}
    \includegraphics[width=\linewidth]{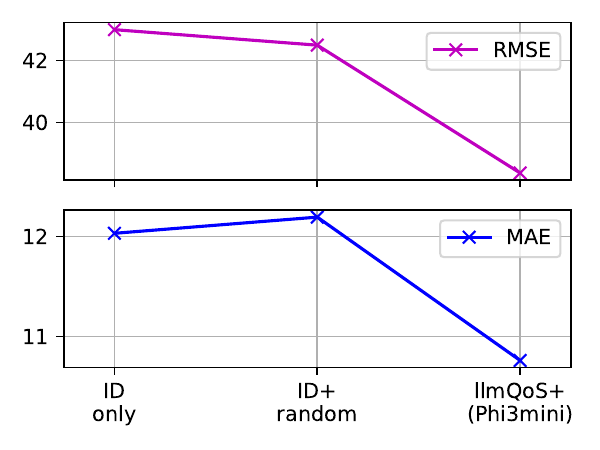}
\end{subfigure}
\begin{subfigure}{0.24\linewidth}
    \caption{Response time $D$=5\%}
    \includegraphics[width=\linewidth]{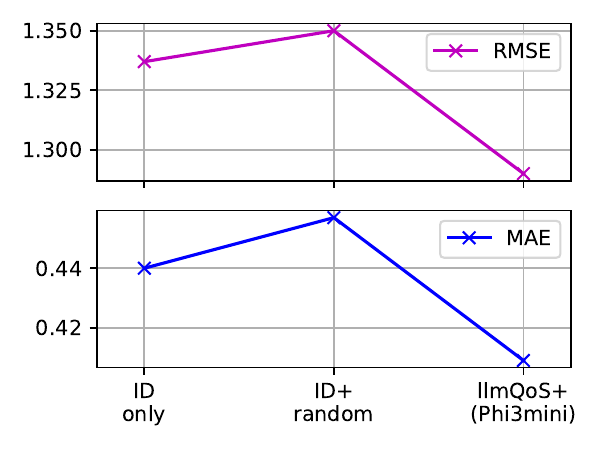}
\end{subfigure}\hfill
\begin{subfigure}{0.24\linewidth}
    \caption{Response time $D$=10\%}
    \includegraphics[width=\linewidth]{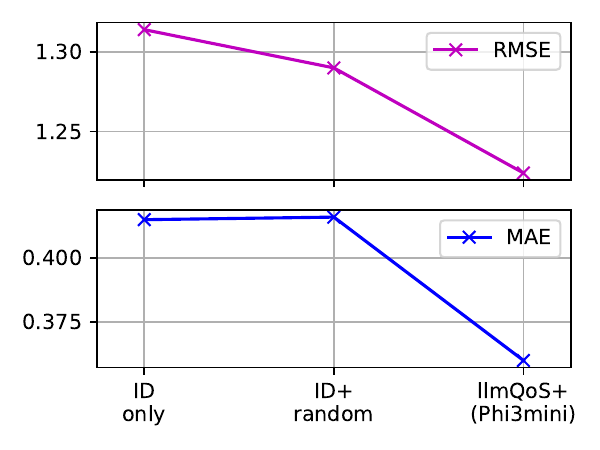}
\end{subfigure}\hfill
\begin{subfigure}{0.24\linewidth}
    \caption{Response time $D$=15\%}
    \includegraphics[width=\linewidth]{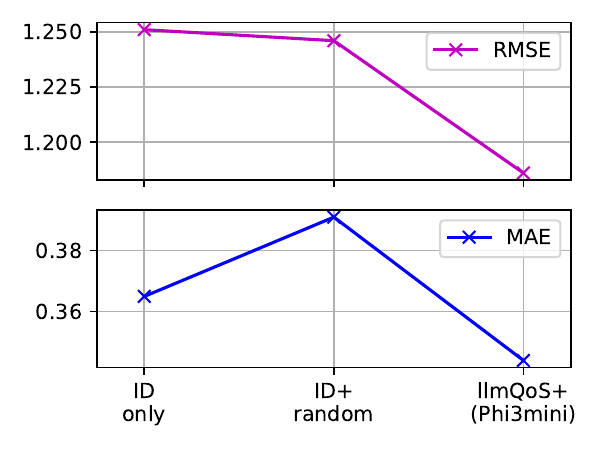}
\end{subfigure}\hfill
\begin{subfigure}{0.24\linewidth}
    \caption{Response time $D$=20\%}
    \includegraphics[width=\linewidth]{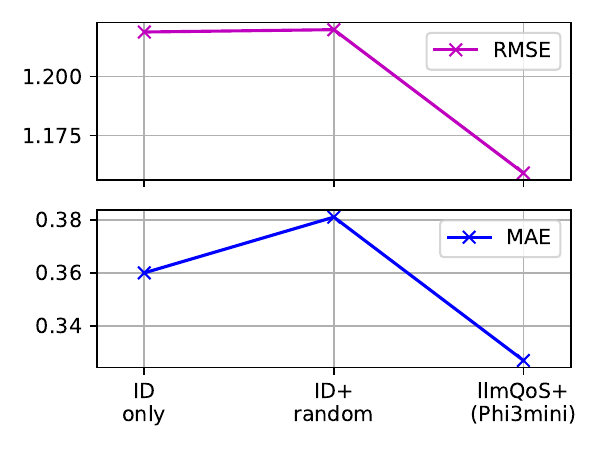}
\end{subfigure}
\caption{QoS prediction performance in terms of Mean Average Error (MAE) and Root Mean Squared Error (RMSE) of model with added random feature vectors as input. Please refer to \autoref{sec:random} for more details. The models with random feature vectors do not outperform the ID only model. This indicates the performance gain obtained by llmQoS is indeed from the useful LLM feature instead of added model parameters.}
\label{fig:random-compare}
\end{figure*}

\begin{figure*}[!ht]\centering
\begin{subfigure}{0.24\linewidth}
    \caption{Throughput $D$=5\%}
    \includegraphics[width=\linewidth]{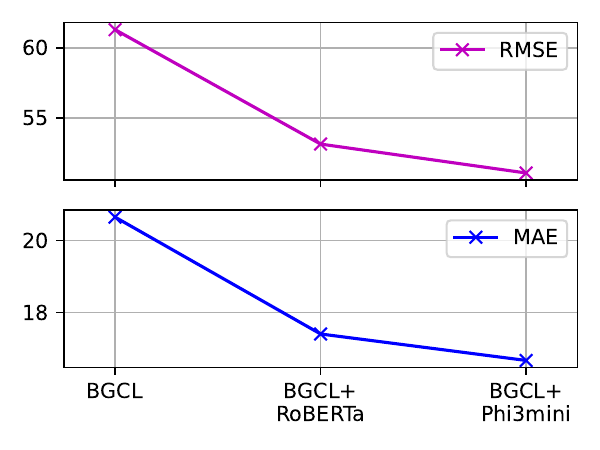}
\end{subfigure}\hfill
\begin{subfigure}{0.24\linewidth}
    \caption{Throughput $D$=10\%}
    \includegraphics[width=\linewidth]{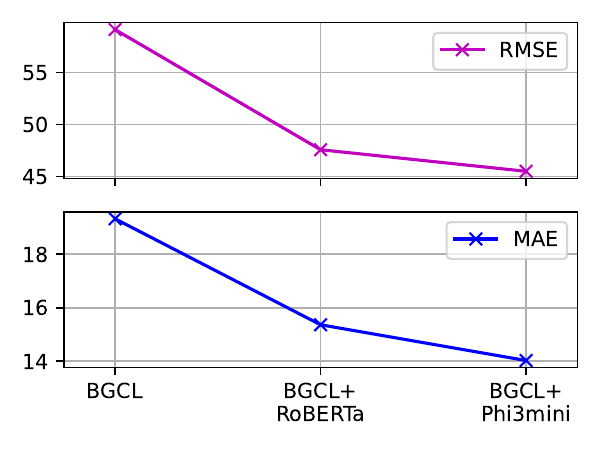}
\end{subfigure}\hfill
\begin{subfigure}{0.24\linewidth}
    \caption{Throughput $D$=15\%}
    \includegraphics[width=\linewidth]{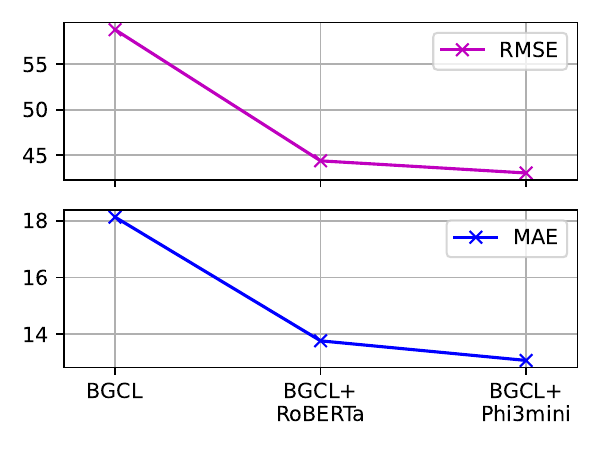}
\end{subfigure}\hfill
\begin{subfigure}{0.24\linewidth}
    \caption{Throughput $D$=20\%}
    \includegraphics[width=\linewidth]{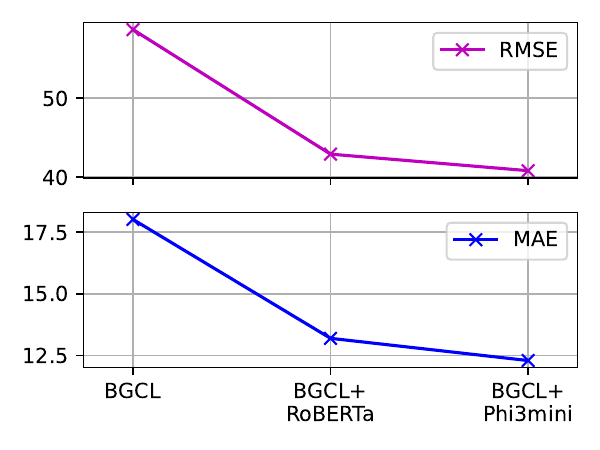}
\end{subfigure}
\begin{subfigure}{0.24\linewidth}
    \caption{Response time $D$=5\%}
    \includegraphics[width=\linewidth]{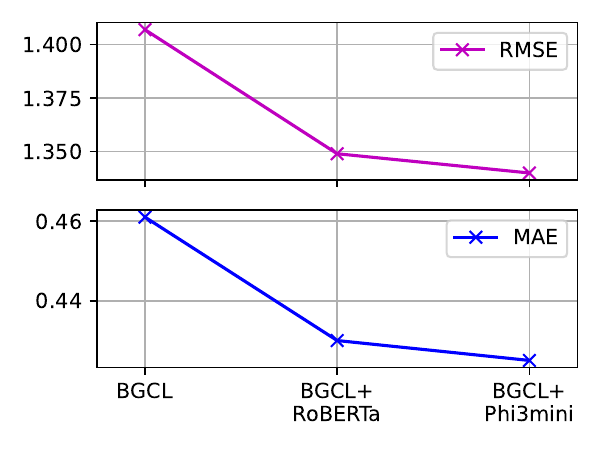}
\end{subfigure}\hfill
\begin{subfigure}{0.24\linewidth}
    \caption{Response time $D$=10\%}
    \includegraphics[width=\linewidth]{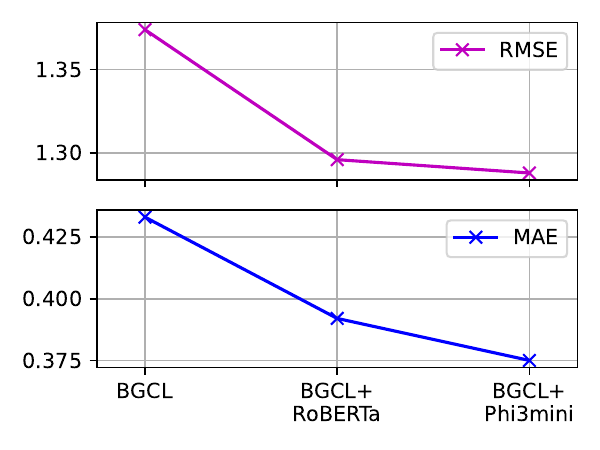}
\end{subfigure}\hfill
\begin{subfigure}{0.24\linewidth}
    \caption{Response time $D$=15\%}
    \includegraphics[width=\linewidth]{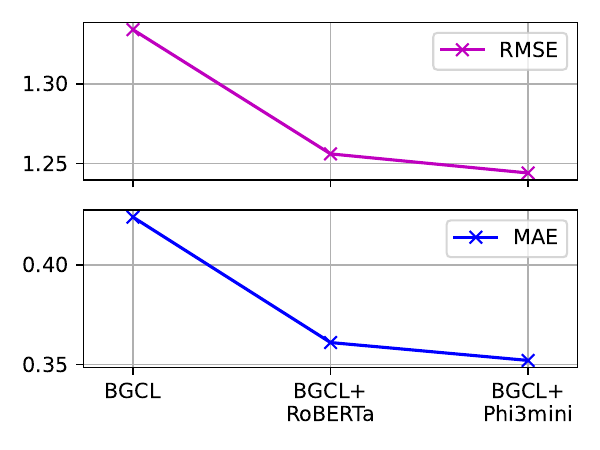}
\end{subfigure}\hfill
\begin{subfigure}{0.24\linewidth}
    \caption{Response time $D$=20\%}
    \includegraphics[width=\linewidth]{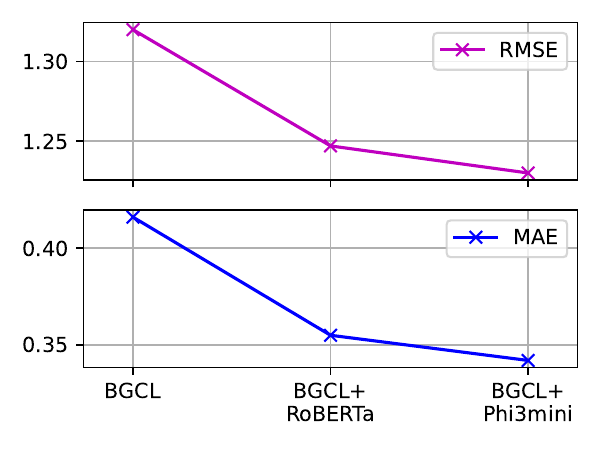}
\end{subfigure}
\caption{QoS prediction performance in terms of Mean Average Error (MAE) and Root Mean Squared Error (RMSE) of BGCL model with different LLM feature extractors. We compare Phi3mini, RoBERTa, and vanilla BGCL. Similar to \autoref{fig:llm-ablation}, LLM features are beneficial for QoS prediction, and while larger Phi3mini achieves better QoS prediction performance than RoBERTa.}
\label{fig:bgcl-compare}
\end{figure*}

\begin{figure*}[!ht]\centering
\begin{subfigure}{0.24\linewidth}
    \caption{Throughput / depth}\label{fig:tune-mlp-depth-tp}
    \includegraphics[width=\linewidth]{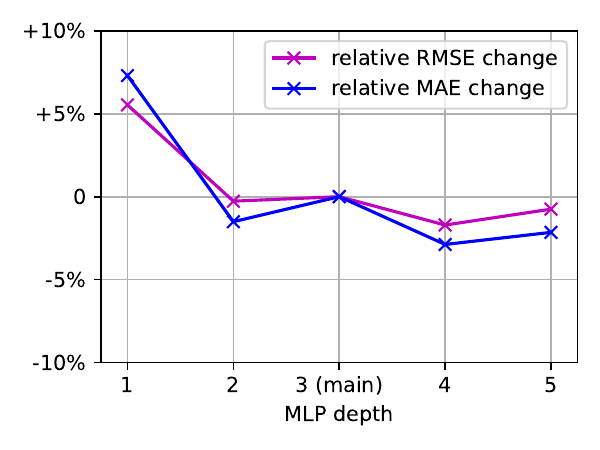}
\end{subfigure}\hfill
\begin{subfigure}{0.24\linewidth}
    \caption{Response / depth}\label{fig:tune-mlp-depth-rt}
    \includegraphics[width=\linewidth]{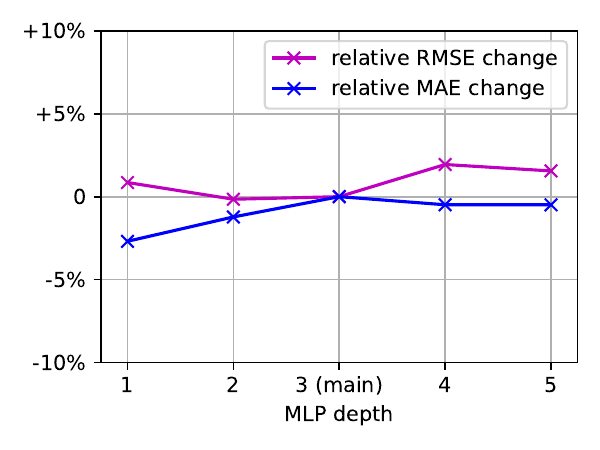}
\end{subfigure}\hfill
\begin{subfigure}{0.24\linewidth}
    \caption{Throughput / width}\label{fig:tune-mlp-width-tp}
    \includegraphics[width=\linewidth]{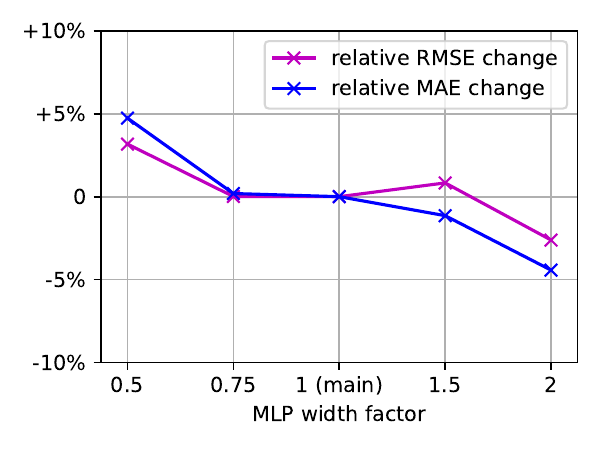}
\end{subfigure}\hfill
\begin{subfigure}{0.24\linewidth}
    \caption{Response time / width}\label{fig:tune-mlp-width-rt}
    \includegraphics[width=\linewidth]{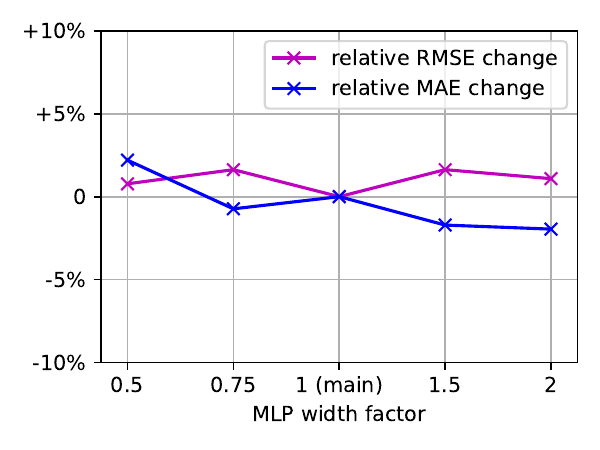}
\end{subfigure}
\caption{QoS prediction performance in terms Mean Average Error (MAE) and Root Mean Squared Error (RMSE) of the proposed llmQoS model different depth and width configurations for the MLP network (\autoref{sec:model-arch}). Depth refers to the number of layers in the MLP. The main experiments use 3 layers. For different width configurations, we multiply the dimensionalities of each layers used by the main experiments by different width factor for comparison. We calculate and show the relative change of MAE and RMSE of a depth or width configuration from the main experiments. This experiment is conducted with density level $D$=5\% and using Phi3mini. Generally, the effect of configuration of MLP has on the QoS prediction performance is insignificant, with changes less than 5\% in most cases.}
\label{fig:tune-mlp}
\end{figure*}

\begin{figure*}[!ht]\centering
\begin{subfigure}{0.24\linewidth}
    \caption{Throughput / BS}\label{fig:tune-bs-tp}
    \includegraphics[width=\linewidth]{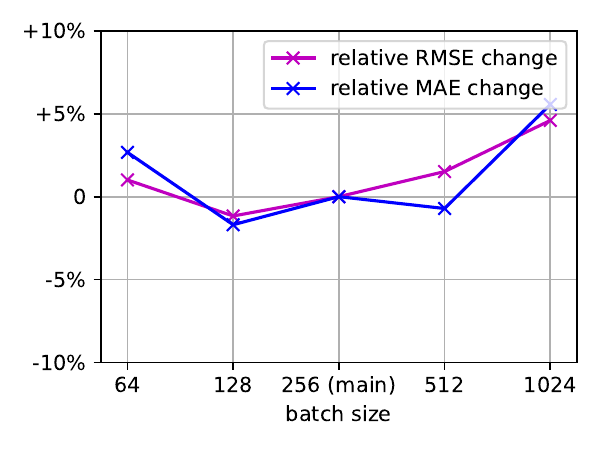}
\end{subfigure}\hfill
\begin{subfigure}{0.24\linewidth}
    \caption{Response time / BS}\label{fig:tune-bs-rt}
    \includegraphics[width=\linewidth]{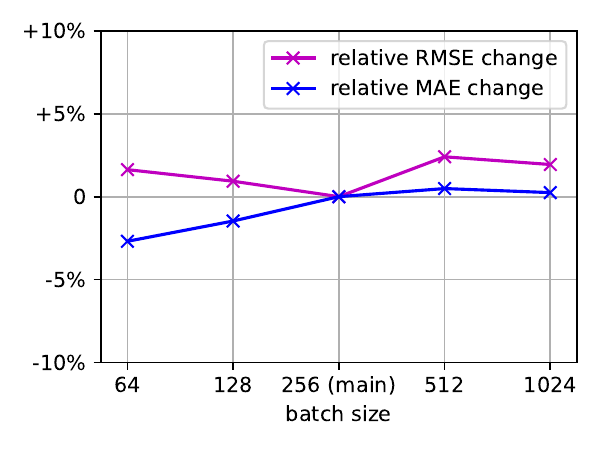}
\end{subfigure}\hfill
\begin{subfigure}{0.24\linewidth} 
    \caption{Throughput / LR}\label{fig:tune-lr-tp}
    \includegraphics[width=\linewidth]{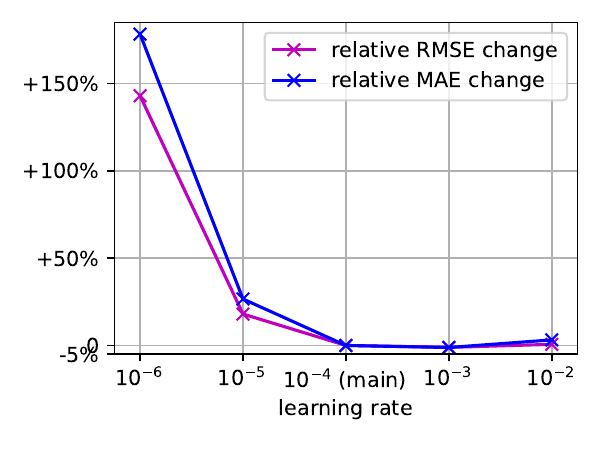}
\end{subfigure}\hfill
\begin{subfigure}{0.24\linewidth}
    \caption{Response time / LR}\label{fig:tune-lr-rt}
    \includegraphics[width=\linewidth]{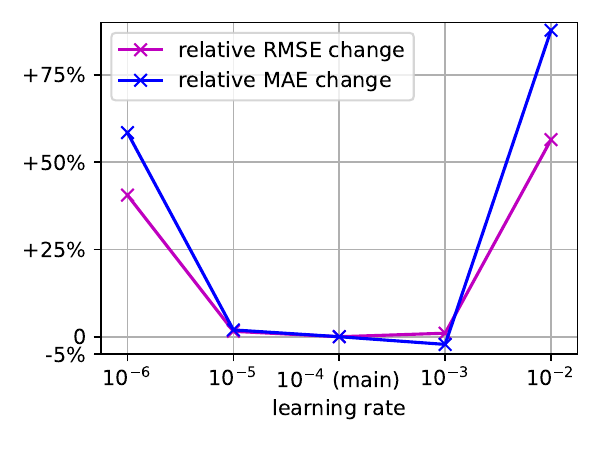}
\end{subfigure}
\caption{QoS prediction performance in terms Mean Average Error (MAE) and Root Mean Squared Error (RMSE) of the proposed llmQoS model trained using different batch sizes (BS) and learning rates (LR). We calculate and show the relative change of MAE and RMSE of a batch size or learning rate from the main experiments. This experiment is conducted with density level $D$=5\% and using Phi3mini. The performance is generally insensitive to batch size. It is sensitive to learning rate, with the optimal value around $10^{-4}$.}
\label{fig:tune-bs-lr}
\end{figure*}

\subsection{Comparison with Baselines}

The performance of the proposed llmQoS model is compared with the baseline methods in \autoref{tab:results}. Phi3mini is used for llmQoS in this experiment. We compare both MAE and RMSE for throughput and response time at different densities. It is evident that llmQoS outperforms all baselines consistently at all densities. For throughput value prediction, llmQoS can reduce the MAE by more than 20\%, and RMSE by more than 6\% from the best baseline model. For response time value prediction, llmQoS can also reduce MAE for more than 10\% from the best baseline method. The prediction accuracy of every method all increases as the QoS matrix density grows. This indicates that with more observed data QoS values, CF-based methods can calculate similarities more accurately, and model-based approaches can capture additional latent features among users and services, leading to better QoS prediction. The results demonstrate that with the rich features from the descriptive sentences extracted by LLMs, the QoS prediction model can effectively find the similarity and correlation among users and services. Along with the information learned from historical interactions, llmQoS can achieve high QoS prediction accuracy.

\subsection{Training Curves}

When setting the training epochs, we configured 1500 epochs for throughput prediction and 600 epochs for response time prediction. We plotted training curves to validate these settings. \autoref{fig:curve} illustrates the loss, MAE, and RMSE curves of the llmQoS model during training at sparse densities ranging from 5\% to 20\%. As the number of iterations increases, the model gradually converges to optimal performance. We observe that the model reaches optimal performance for throughput prediction before 1500 epochs, while for response time prediction, it reaches optimal performance before 600 epochs.

\subsection{Comparison of Different LLMs}\label{sec:compare-roberta-phi3}

As discussed in \autoref{sec:related-llm} and \autoref{sec:llm-extraction}, larger LLMs have higher capacity and can potentially achieve higher performance on various tasks. So here we compare the performance of llmQoS with different LLMs as the feature extractor. We compare the larger and more recent Phi3mini with the smaller and older RoBERTa model. In addition, we remove all LLM feature related components from llmQoS, and only use the ID embedding vectors of users and services to train a QoS prediction model. This model is referred as ``ID only'' model. We train all models using the same training hyper-parameters, and the results are compared in \autoref{fig:llm-ablation}.

It is clear that adding LLM features to QoS prediction is helpful, and can reduce the prediction error significantly from the ID only baseline. This further verifies that the information from the attributes of users and services is useful for service recommendation. And Phi3mini achieves lower MAE and RMSE than RoBERTa consistently on all densities. This shows that the superior capability of larger LLMs on NLP tasks also transfers to QoS prediction task.

However, it is important to note that larger LLMs runs at a higher computational cost. Phi3mini is considered one of the smallest modern LLMs, as the larger ones have tens or hundreds of billions of parameters. Larger LLMs can also have higher feature dimensionality, making training the QoS prediction network slower. In practice, trade-off needs to be made between prediction accuracy and computational cost.



\subsection{Ablation Study Using Random Feature}\label{sec:random}

In \autoref{sec:compare-roberta-phi3} we have shown that LLM feature improves QoS prediction accuracy, and larger Phi3mini is more effective than smaller RoBERTa. However, the prediction model with LLM feature input has more trainable parameters than the model that only uses ID embedding. What is more, since the feature vectors from Phi3mini have 3,072 dimensions compared to RoBERTa's 768 dimensions, the projection layer for Phi3mini also has more parameters than RoBERTa. So it is plausible that the performance improvement is due to the increased number of parameters resulting in large model capacity.

To verify this, we train a model that still has 3,072-dimensional feature vectors as input. But those feature vectors are randomly initialized for each user and service, so they do not contain any useful information. However, this model has the same number of parameters as the llmQoS model with Phi3mini feature. We compare the QoS prediction performance of this model with both the ID only model as in \autoref{sec:compare-roberta-phi3} and the llmQoS with Phi3mini in \autoref{fig:random-compare}. Generally, the model with random feature does not outperform ID only model. On throughput dataset under $D$=5\% and $D$=15\%, the accuracy is marginally higher than the ID only model. But under other settings the performance actually worsens. We also observe that with random feature, the model converges with much less epochs than using Phi3mini feature. This indicates that more trainable parameters lead to overfitting. We verify that the performance boost of llmQoS is from the useful information encoded in the LLM feature, instead of added parameters of the model.

\subsection{Adding LLM Feature to BGCL}\label{sec:bgcl}

To if LLM features can benefit QoS prediction in general, we further conduct experiments by incorporating LLM features into another QoS prediction model. We selected the Bi-subgraph network based on graph contrastive learning (BGCL) model \cite{BGCL}, which has a more complex network architecture. BGCL uses location attributes of users and services for graph contrastive learning, as well as attention mechanism to aggregate features. If LLM features can improve the QoS prediction accuracy for BGCL, we claim that the benefits of using LLM can be further demonstrated.

We keep all other features and network architecture unchanged for BGCL, with the only addition of the LLM feature vectors along side with the embedding vectors of BGCL at the very beginning of the network of BGCL. Similar to \autoref{sec:compare-roberta-phi3}, we compare BGCL with Phi3mini, BGCL with RoBERTa, and vanilla BGCL. The experimental results, as shown in \autoref{fig:bgcl-compare}, are similar to the results in \autoref{fig:llm-ablation} too. We further verify that that integrating LLM features significantly enhances the model's performance, and larger model achieves more satisfactory results. This demonstrates that LLM features are beneficial to QoS prediction problem, regardless of the network architecture. And we expect this to apply to other QoS prediction models as well.




\subsection{Hyper-Parameter Tuning of MLP}

We test the performance of llmQoS using different configurations for the MLP model as described in \autoref{sec:model-arch}. For this experiment we test the density level $D$=5\%, and using Phi3mini as the LLM feature extractor. We vary the depth of the MLP by adding or removing layers from it. The QoS prediction performance of different configurations is shown in \autoref{fig:tune-mlp-depth-tp} and \ref{fig:tune-mlp-depth-rt}. We change the width of the MLP by multiplying a width factor to the dimensionalities of the layers in it. The QoS prediction performance of different configurations is shown in \autoref{fig:tune-mlp-width-tp} and \ref{fig:tune-mlp-width-rt}.

It can be observed that the depth and width of the MLP network have a minor impact on the QoS prediction performance. In most cases the change in MAE and RMSE is under 5\%. However, when the MLP depth is set to 1, or the width factor is set to 0.5, there is a significant deterioration in the performance. This is caused by the network is too shallow and narrow, resulting to very low capacity. Considering this paper primarily focuses on using large language models to assist QoS prediction, we have selected a reasonably appropriate depth and width for the MLP network without extensive searching.

\subsection{Training Parameter Tuning}

We test the performance of llmQoS trained using different batch sizes and learning rates. For this experiment we test the density level $D$=5\%, and using Phi3mini as the LLM feature extractor. The same configuration of MLP used in the main experiments is used.

The QoS performance of models trained using different batch sizes is compared in \autoref{fig:tune-bs-tp} and \ref{fig:tune-bs-rt}. Generally speaking, the impact of batch size is inconsistent and insignificant. Only for throughput, larger batch size of 1024 causes the prediction accuracy to drop significantly. The performance with different learning rates is compared in \autoref{fig:tune-lr-tp} and \ref{fig:tune-lr-rt}. Learning rate has a huge impact on the model's prediction accuracy. Very small learning rate results in slow convergence, while large learning rate causes unstable training. Both causes low performance. The optimal learning rate is around $10^{-4}$ as used in the main experiments.

\subsection{Limitations}

In various experiments, we have shown that the proposed llmQoS model has promising QoS prediction capabilities. However, our study has limitations. The straightforward MLP architecture of the llmQoS model, while demonstrating effectiveness, may not capture all deeply complex user-service interaction patterns. The computational overhead associated with LLMs remains also a practical consideration. The impact on QoS prediction performance of the specific methodology for constructing descriptive sentences from attributes is also not fully explored. The inherent characteristics or pre-training nuances of the selected LLMs could also influence the quality of extracted features and, consequently, the prediction performance.

\section{Conclusion}

In this paper, we have proposed the large language model aided QoS prediction (llmQoS) model. It utilizes LLMs to extract features from descriptive sentences constructed based on attributes of web users and services. Then the LLM features are combined with ID based embedding features for QoS prediction through a multi-layer perceptron network. Despite its concise network architecture, llmQoS achieves accurate QoS prediction under data sparsity condition, and outperforms comparable baseline models consistently on the WSDream dataset. We have also shown that the benefits of adding LLM features is general, and larger LLMs achieve better performance. In the era of explosive development of LLMs, we hope our paper can inspire more researchers in the field of service recommendation to conduct research in this direction.

\bibliographystyle{IEEEtran}
\bibliography{refs}

\begin{thebibliography}{10}
\providecommand{\url}[1]{#1}
\csname url@samestyle\endcsname
\providecommand{\newblock}{\relax}
\providecommand{\bibinfo}[2]{#2}
\providecommand{\BIBentrySTDinterwordspacing}{\spaceskip=0pt\relax}
\providecommand{\BIBentryALTinterwordstretchfactor}{4}
\providecommand{\BIBentryALTinterwordspacing}{\spaceskip=\fontdimen2\font plus
\BIBentryALTinterwordstretchfactor\fontdimen3\font minus
  \fontdimen4\font\relax}
\providecommand{\BIBforeignlanguage}[2]{{%
\expandafter\ifx\csname l@#1\endcsname\relax
\typeout{** WARNING: IEEEtran.bst: No hyphenation pattern has been}%
\typeout{** loaded for the language `#1'. Using the pattern for}%
\typeout{** the default language instead.}%
\else
\language=\csname l@#1\endcsname
\fi
#2}}
\providecommand{\BIBdecl}{\relax}
\BIBdecl

\bibitem{introduction1}
S.~Li, H.~Luo, and G.~Zhao, ``bi-{HPTM}: An effective semantic matchmaking
  model for web service discovery,'' in \emph{IEEE International Conference on
  Web Services}, 2020.

\bibitem{introduction2}
M.~Silic, G.~Delac, and S.~Srbljic, ``Prediction of atomic web services
  reliability based on k-means clustering,'' in \emph{Joint Meeting on
  Foundations of Software Engineering}, 2013.

\bibitem{SSE2}
F.~Sun, G.~Chen, H.~Ma, and S.~Hartmann, ``Population-based incremental
  learning for effective iot service composition with replication,'' in
  \emph{IEEE International Conference on Software Services Engineering (SSE)},
  2024.

\bibitem{SSE5}
S.~Baravkar, C.~Zhang, F.~Hassan, L.~Cheng, and Z.~Song, ``Decoding and
  answering developers' questions about web services managed by marketplaces,''
  in \emph{IEEE International Conference on Software Services Engineering
  (SSE)}, 2024.

\bibitem{introduction3}
K.~Qi, H.~Hu, W.~Song, J.~Ge, and J.~Lü, ``Personalized qos prediction via
  matrix factorization integrated with neighborhood information,'' in
  \emph{IEEE International Conference on Services Computing}, 2015.

\bibitem{SSE4}
H.~Ge, Q.~Li, S.~Meng, and J.~Hou, ``Cpgcn: Collaborative property-aware graph
  convolutional networks for service recommendation,'' in \emph{IEEE
  International Conference on Services Computing (SCC)}, 2022.

\bibitem{Zhao2023ASO}
W.~X. Zhao, K.~Zhou, J.~Li, T.~Tang, X.~Wang, Y.~Hou, Y.~Min, B.~Zhang,
  J.~Zhang, Z.~Dong, Y.~Du, C.~Yang, Y.~Chen, Z.~Chen, J.~Jiang, R.~Ren, Y.~Li,
  X.~Tang, Z.~Liu, P.~Liu, J.~Nie, and J.~rong Wen, ``A survey of large
  language models,'' \emph{ArXiv}, 2023.

\bibitem{Minaee2024LargeLM}
S.~Minaee, T.~Mikolov, N.~Nikzad, M.~A. Chenaghlu, R.~Socher, X.~Amatriain, and
  J.~Gao, ``Large language models: A survey,'' \emph{ArXiv}, 2024.

\bibitem{devlin2019bert}
J.~Devlin, M.-W. Chang, K.~Lee, and K.~Toutanova, ``{BERT}: Pre-training of
  deep bidirectional transformers for language understanding,'' in
  \emph{Conference of the North {A}merican Chapter of the Association for
  Computational Linguistics: Human Language Technologies}, 2019.

\bibitem{liu2019roberta}
Y.~Liu, M.~Ott, N.~Goyal, J.~Du, M.~Joshi, D.~Chen, O.~Levy, M.~Lewis,
  L.~Zettlemoyer, and V.~Stoyanov, ``Roberta: A robustly optimized bert
  pretraining approach,'' \emph{ArXiv}, 2019.

\bibitem{brown2020languagemodelsfewshotlearners}
T.~B. Brown, B.~Mann, N.~Ryder, M.~Subbiah, and et~al., ``Language models are
  few-shot learners,'' \emph{OpenAI}, 2020.

\bibitem{abdin2024phi3}
M.~Abdin, S.~A. Jacobs, A.~A. Awan, J.~Aneja, and et~al., ``{Phi}-3 technical
  report: A highly capable language model locally on your phone,''
  \emph{ArXiv}, 2024.

\bibitem{meta2024llama3}
\BIBentryALTinterwordspacing
{Meta AI}, ``Introducing meta {Llama} 3: The most capable openly available llm
  to date,'' 2024. [Online]. Available:
  \url{https://ai.meta.com/blog/meta-llama-3/}
\BIBentrySTDinterwordspacing

\bibitem{CF}
B.~Sarwar, G.~Karypis, J.~Konstan, and J.~Riedl, ``Item-based collaborative
  filtering recommendation algorithms,'' in \emph{International Conference on
  World Wide Web}, 2001.

\bibitem{user-based1}
L.~Shao, J.~Zhang, Y.~Wei, J.~Zhao, B.~Xie, and H.~Mei, ``Personalized qos
  prediction forweb services via collaborative filtering,'' in \emph{IEEE
  International Conference on Web Services}, 2007.

\bibitem{user-based-cold}
L.~Qi, W.~Dou, and X.~Zhang, ``An inverse collaborative filtering approach for
  cold-start problem in web service recommendation,'' in \emph{Australasian
  Computer Science Week Multiconference}, 2017.

\bibitem{item-based1}
B.~Sarwar, G.~Karypis, J.~Konstan, and J.~Riedl, ``Item-based collaborative
  filtering recommendation algorithms,'' in \emph{International Conference on
  World Wide Web}, 2001.

\bibitem{item-based2}
G.~Linden, B.~Smith, and J.~York, ``Amazon.com recommendations: item-to-item
  collaborative filtering,'' \emph{IEEE Internet Computing}, vol.~7, no.~1, pp.
  76--80, 2003.

\bibitem{item-based-loc}
Z.~Chen, L.~Shen, and F.~Li, ``Exploiting web service geographical neighborhood
  for collaborative qos prediction,'' \emph{Future Generation Computer
  Systems}, vol.~68, pp. 248--259, 2017.

\bibitem{item-user-based1}
Y.~Ma, S.~Wang, P.~C. Hung, C.-H. Hsu, Q.~Sun, and F.~Yang, ``A highly accurate
  prediction algorithm for unknown web service qos values,'' \emph{IEEE
  Transactions on Services Computing}, vol.~9, no.~4, pp. 511--523, 2016.

\bibitem{UIPCC}
Z.~Zheng, H.~Ma, M.~R. Lyu, and I.~King, ``Qos-aware web service recommendation
  by collaborative filtering,'' \emph{IEEE Transactions on Services Computing},
  vol.~4, no.~2, pp. 140--152, 2011.

\bibitem{time_and_sim}
Y.~Hu, Q.~Peng, X.~Hu, and R.~Yang, ``Time aware and data sparsity tolerant web
  service recommendation based on improved collaborative filtering,''
  \emph{IEEE Transactions on Services Computing}, vol.~8, no.~5, pp. 782--794,
  2015.

\bibitem{AMF}
J.~Zhu, P.~He, Z.~Zheng, and M.~R. Lyu, ``Online qos prediction for runtime
  service adaptation via adaptive matrix factorization,'' \emph{IEEE
  Transactions on Parallel and Distributed Systems}, vol.~28, no.~10, pp.
  2911--2924, 2017.

\bibitem{NIMF}
Z.~Zheng, H.~Ma, M.~R. Lyu, and I.~King, ``Collaborative web service qos
  prediction via neighborhood integrated matrix factorization,'' \emph{IEEE
  Transactions on Services Computing}, vol.~6, no.~3, pp. 289--299, 2013.

\bibitem{PMF-QoS}
Y.~Xu, J.~Yin, W.~Lo, and Z.~Wu, ``Personalized location-aware qos prediction
  for web services using probabilistic matrix factorization,'' in \emph{Web
  Information Systems Engineering}, X.~Lin, Y.~Manolopoulos, D.~Srivastava, and
  G.~Huang, Eds., 2013.

\bibitem{Silver2016AphaGo}
D.~Silver, A.~Huang, C.~J. Maddison, A.~Guez, L.~Sifre, G.~van~den Driessche,
  J.~Schrittwieser, I.~Antonoglou, V.~Panneershelvam, M.~Lanctot, S.~Dieleman,
  D.~Grewe, J.~Nham, N.~Kalchbrenner, I.~Sutskever, T.~Lillicrap, M.~Leach,
  K.~Kavukcuoglu, T.~Graepel, and D.~Hassabis, ``Mastering the game of {Go}
  with deep neural networks and tree search,'' \emph{Nature}, vol. 529, no.
  7587, pp. 484--489, 2016.

\bibitem{Chai2021DLCV}
J.~Chai, H.~Zeng, A.~Li, and E.~W. Ngai, ``Deep learning in computer vision: A
  critical review of emerging techniques and application scenarios,''
  \emph{Machine Learning with Applications}, vol.~6, p. 100134, 2021.

\bibitem{Soori2023DLRobotics}
M.~Soori, B.~Arezoo, and R.~Dastres, ``Artificial intelligence, machine
  learning and deep learning in advanced robotics, a review,'' \emph{Cognitive
  Robotics}, vol.~3, pp. 54--70, 2023.

\bibitem{SSE1}
Y.~Zeng, Y.~Li, Z.~Xia, Z.~Du, J.~Wang, R.~Lian, and J.~Xu, ``Qoseraser: A data
  erasable framework for web service qos prediction,'' in \emph{IEEE
  International Conference on Software Services Engineering (SSE)}, 2023.

\bibitem{deeplearning1}
C.~Wei, Y.~Fan, and J.~Zhang, ``Time-aware service recommendation with
  social-powered graph hierarchical attention network,'' \emph{IEEE
  Transactions on Services Computing}, vol.~16, no.~3, pp. 2229--2240, 2023.

\bibitem{deeplearning2}
Y.~Yin, Q.~Di, J.~Wan, and T.~Liang, ``Time-aware smart city services based on
  qos prediction: A contrastive learning approach,'' \emph{IEEE Internet of
  Things Journal}, vol.~10, no.~21, pp. 18\,745--18\,753, 2023.

\bibitem{SSE6}
B.~Hu, X.~Xie, J.~Shen, J.~Zhang, S.~J. Lee, and S.~Lee, ``High-order-modal
  knowledge graph powered api recommendation for mashup development,'' in
  \emph{IEEE International Conference on Software Services Engineering (SSE)},
  2024.

\bibitem{NCRL}
G.~Zou, S.~Wu, S.~Hu, C.~Cao, Y.~Gan, B.~Zhang, and Y.~Chen, ``Ncrl:
  Neighborhood-based collaborative residual learning for adaptive qos
  prediction,'' \emph{IEEE Transactions on Services Computing}, vol.~16, no.~3,
  pp. 2030--2043, 2023.

\bibitem{Robust}
Z.~Wu, D.~Ding, Y.~Xiu, Y.~Zhao, and J.~Hong, ``Robust qos prediction based on
  reputation integrated graph convolution network,'' \emph{IEEE Transactions on
  Services Computing}, vol.~17, no.~3, pp. 1154--1167, 2024.

\bibitem{DRGL}
C.~Wei, Y.~Fan, J.~Zhang, Z.~Jia, and R.~Yan, ``Dynamic relation graph learning
  for time-aware service recommendation,'' \emph{IEEE Transactions on Network
  and Service Management}, vol.~21, no.~2, pp. 1503--1517, 2024.

\bibitem{BGCL}
J.~Zhu, B.~Li, J.~Wang, D.~Li, Y.~Liu, and Z.~Zhang, ``Bgcl: Bi-subgraph
  network based on graph contrastive learning for cold-start qos prediction,''
  \emph{Knowledge-Based Systems}, vol. 263, p. 110296, 2023.

\bibitem{SCATSF}
J.~Zhou, D.~Ding, Z.~Wu, and Y.~Xiu, ``Spatial context-aware time-series
  forecasting for {QoS} prediction,'' \emph{IEEE Transactions on Network and
  Service Management}, vol.~20, no.~2, pp. 918--931, 2023.

\bibitem{PMT}
H.~Lian, J.~Li, H.~Wu, Y.~Zhao, L.~Zhang, and X.~Wang, ``Toward effective
  personalized service {QoS} prediction from the perspective of multi-task
  learning,'' \emph{IEEE Transactions on Network and Service Management},
  vol.~20, no.~3, pp. 2587--2597, 2023.

\bibitem{NIPS2017Transformer}
A.~Vaswani, N.~Shazeer, N.~Parmar, J.~Uszkoreit, L.~Jones, A.~N. Gomez, L.~u.
  Kaiser, and I.~Polosukhin, ``Attention is all you need,'' in \emph{Advances
  in Neural Information Processing Systems}, 2017.

\bibitem{lan2020albert}
Z.~Lan, M.~Chen, S.~Goodman, K.~Gimpel, P.~Sharma, and R.~Soricut, ``{ALBERT:}
  {A} lite {BERT} for self-supervised learning of language representations,''
  in \emph{International Conference on Learning Representations}, 2020.

\bibitem{Radford2018ImprovingLU}
A.~Radford and K.~Narasimhan, ``Improving language understanding by generative
  pre-training,'' \emph{OpenAI}, 2018.

\bibitem{solaiman2019release}
I.~Solaiman, M.~Brundage, J.~Clark, A.~Askell, A.~Herbert-Voss, J.~Wu,
  A.~Radford, G.~Krueger, J.~W. Kim, S.~Kreps, M.~McCain, A.~Newhouse,
  J.~Blazakis, K.~McGuffie, and J.~Wang, ``Release strategies and the social
  impacts of language models,'' \emph{ArXiv}, 2019.

\bibitem{openai2024gpt4technicalreport}
J.~Achiam, S.~Adler, S.~Agarwal, L.~Ahmad, and et~al., ``{GPT}-4 technical
  report,'' \emph{OpenAI}, 2024.

\bibitem{Touvron2023LLaMAOA}
H.~Touvron, T.~Lavril, G.~Izacard, X.~Martinet, M.-A. Lachaux, T.~Lacroix,
  B.~Rozi{\`e}re, N.~Goyal, E.~Hambro, F.~Azhar, A.~Rodriguez, A.~Joulin,
  E.~Grave, and G.~Lample, ``{LLaMA}: Open and efficient foundation language
  models,'' \emph{ArXiv}, 2023.

\bibitem{Touvron2023Llama2O}
H.~Touvron, L.~Martin, K.~R. Stone, P.~Albert, and et~al., ``{Llama} 2: Open
  foundation and fine-tuned chat models,'' \emph{ArXiv}, 2023.

\bibitem{gunasekar2023textbooksneed}
S.~Gunasekar, Y.~Zhang, J.~Aneja, C.~C.~T. Mendes, and et~al., ``Textbooks are
  all you need,'' \emph{ArXiv}, 2023.

\bibitem{li2023textbooksneediiphi15}
Y.~Li, S.~Bubeck, R.~Eldan, A.~D. Giorno, S.~Gunasekar, and Y.~T. Lee,
  ``Textbooks are all you need ii: phi-1.5 technical report,'' \emph{ArXiv},
  2023.

\bibitem{ms2023phi2}
\BIBentryALTinterwordspacing
M.~Abdin, J.~Aneja, S.~Bubeck, C.~C.~T. Mendes, and et~al., ``{Phi}-2: The
  surprising power of small language models,'' 2024. [Online]. Available:
  \url{https://www.microsoft.com/en-us/research/blog/phi-2-the-surprising-power-of-small-language-models/}
\BIBentrySTDinterwordspacing

\bibitem{HFOpenLLMLeaderboard}
\BIBentryALTinterwordspacing
{HuggingFace}, ``Open {LLM} leaderboard,'' 2024. [Online]. Available:
  \url{https://huggingface.co/spaces/open-llm-leaderboard/open_llm_leaderboard}
\BIBentrySTDinterwordspacing

\bibitem{Guo2022FromIT}
J.~Guo, J.~Li, D.~Li, A.~M.~H. Tiong, B.~A. Li, D.~Tao, and S.~C.~H. Hoi,
  ``From images to textual prompts: Zero-shot visual question answering with
  frozen large language models,'' in \emph{IEEE/CVF Conference on Computer
  Vision and Pattern Recognition}, 2022.

\bibitem{liu2023grounding}
S.~Liu, Z.~Zeng, T.~Ren, F.~Li, H.~Zhang, J.~Yang, C.~Li, J.~Yang, H.~Su,
  J.~Zhu \emph{et~al.}, ``Grounding dino: Marrying dino with grounded
  pre-training for open-set object detection,'' \emph{ArXiv}, 2023.

\bibitem{song2023llmplanner}
C.~H. Song, J.~Wu, C.~Washington, B.~M. Sadler, W.-L. Chao, and Y.~Su,
  ``{LLM}-{Planner}: Few-shot grounded planning for embodied agents with large
  language models,'' in \emph{IEEE/CVF International Conference on Computer
  Vision}, 2023.

\bibitem{fu2023drive}
D.~Fu, X.~Li, L.~Wen, M.~Dou, P.~Cai, B.~Shi, and Y.~Qiao, ``Drive like a
  human: Rethinking autonomous driving with large language models,''
  \emph{ArXiv}, 2023.

\bibitem{wsdream2008}
Z.~Zheng and M.~R. Lyu, ``Ws-dream: A distributed reliability assessment
  mechanism for web services,'' in \emph{IEEE International Conference on
  Dependable Systems and Networks}, 2008.

\bibitem{chollet2015keras}
F.~Chollet \emph{et~al.}, ``Keras,'' \url{https://keras.io}, 2015.

\bibitem{tensorflow2015whitepaper}
\BIBentryALTinterwordspacing
M.~Abadi, A.~Agarwal, P.~Barham, E.~Brevdo, and et~al., ``{TensorFlow}:
  Large-scale machine learning on heterogeneous systems,'' 2015, software
  available from tensorflow.org. [Online]. Available:
  \url{https://www.tensorflow.org/}
\BIBentrySTDinterwordspacing

\bibitem{huggingface}
T.~Wolf, L.~Debut, V.~Sanh, J.~Chaumond, C.~Delangue, A.~Moi, P.~Cistac,
  T.~Rault, R.~Louf, M.~Funtowicz, and J.~Brew, ``Huggingface's transformers:
  State-of-the-art natural language processing,'' \emph{ArXiv}, 2019.

\bibitem{agarap2018deep}
A.~F. Agarap, ``Deep learning using rectified linear units (relu),''
  \emph{ArXiv}, 2018.

\bibitem{Huber1964RobustEO}
P.~J. Huber, ``Robust estimation of a location parameter,'' \emph{Annals of
  Mathematical Statistics}, vol.~35, pp. 492--518, 1964.

\bibitem{KingmaB14Adam}
D.~P. Kingma and J.~Ba, ``Adam: {A} method for stochastic optimization,'' in
  \emph{International Conference on Learning Representations}, 2015.

\bibitem{RegionKNN}
X.~Chen, X.~Liu, Z.~Huang, and H.~Sun, ``Regionknn: A scalable hybrid
  collaborative filtering algorithm for personalized web service
  recommendation,'' in \emph{IEEE International Conference on Web Services},
  2010.

\bibitem{LACF}
M.~Tang, Y.~Jiang, J.~Liu, and X.~Liu, ``Location-aware collaborative filtering
  for qos-based service recommendation,'' in \emph{IEEE International
  Conference on Web Services}, 2012.

\bibitem{PMF}
R.~Salakhutdinov and A.~Mnih, ``Probabilistic matrix factorization,'' in
  \emph{International Conference on Neural Information Processing Systems},
  2007.

\bibitem{PSO-USRec}
J.~Chen, C.~Mao, and W.~W. Song, ``Qos prediction for web services in cloud
  environments based on swarm intelligence search,'' \emph{Knowledge-Based
  Systems}, vol. 259, p. 110081, 2023.

\bibitem{LMF-PP}
D.~Ryu, K.~Lee, and J.~Baik, ``Location-based web service qos prediction via
  preference propagation to address cold start problem,'' \emph{IEEE
  Transactions on Services Computing}, vol.~14, no.~3, pp. 736--746, 2021.

\bibitem{DCALF}
D.~Wu, X.~Luo, M.~Shang, Y.~He, G.~Wang, and X.~Wu, ``A
  data-characteristic-aware latent factor model for web services qos
  prediction,'' \emph{IEEE Transactions on Knowledge and Data Engineering},
  vol.~34, no.~6, pp. 2525--2538, 2022.

\end{thebibliography}

\end{document}